\documentclass[runningheads]{llncs}

\usepackage{graphicx}
\usepackage{comment}
\usepackage{amsmath,amssymb}
\DeclareMathOperator*{\argmax}{arg\,max}
\usepackage{color}
\usepackage{url}
\usepackage{hyperref}
\usepackage{times}
\usepackage{epsfig}
\usepackage{subcaption}
\usepackage{booktabs}
\usepackage{multicol}
\usepackage{multirow}
\usepackage{sidecap}
\usepackage[justification=justified,format=plain,size=small]{caption}

\newif\ifreview
\reviewfalse

\ifreview
	\usepackage{lineno}

	\linenumbers
\fi

\begin{document}


\def\SubNumber{105}

\def\GCPRTrack{Fast Review Track}

\title{Multiclass Alignment of Confidence and Certainty for Network Calibration}

\ifreview
\else

	
	
        \author{Vinith Kugathasan\and
	Muhammad Haris Khan}
	
	\authorrunning{Vinith and Muhammad Haris et al.}
	
	\institute{Mohamed bin Zayed University of Artificial Intelligence, UAE
        \email{\{muhammad.haris\}@mbzuai.ac.ae}}
\fi

\maketitle              

\begin{abstract}
Deep neural networks (DNNs) have made great strides in pushing the state-of-the-art in several challenging domains. Recent studies reveal that they are prone to making overconfident predictions. This greatly reduces the overall trust in model predictions, especially in safety-critical applications. Early work in improving model calibration employs post-processing techniques which rely on limited parameters and require a hold-out set. Some recent train-time calibration methods, which involve all model parameters, can outperform the post-processing methods. To this end, we propose a new train-time calibration method, which features a simple, plug-and-play auxiliary loss known as multi-class alignment of predictive mean confidence and predictive certainty (MACC). It is based on the observation that a model miscalibration is directly related to its predictive certainty, so a higher gap between the mean confidence and certainty amounts to a poor calibration both for in-distribution and out-of-distribution predictions. Armed with this insight, our proposed loss explicitly encourages a confident (or underconfident) model to also provide a low (or high) spread in the pre-softmax distribution. Extensive experiments on ten challenging datasets, covering in-domain, out-domain, non-visual recognition and medical image classification scenarios, show that our method achieves state-of-the-art calibration performance for both in-domain and out-domain predictions. Our code and models will be publicly released.

\end{abstract}
\begin{figure*}[ht]
     \centering
     \begin{subfigure}[b]{0.49\linewidth}
         \centering
         \includegraphics[width=\linewidth]{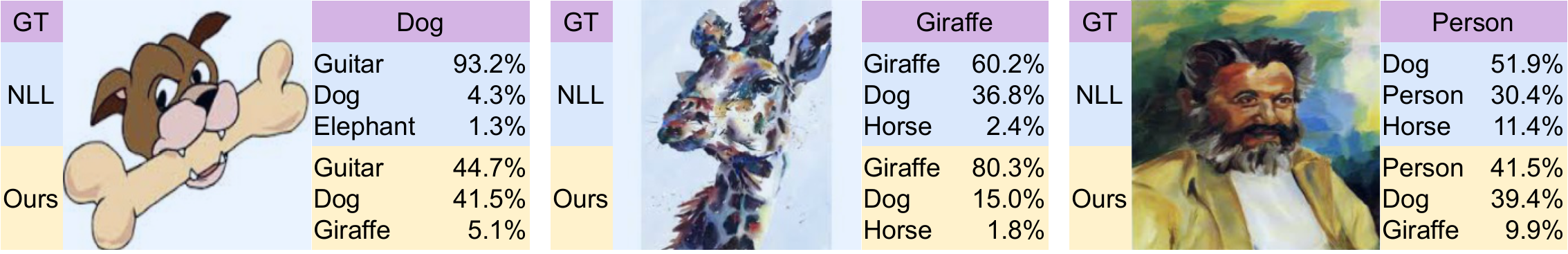}
         \vspace{-0.6cm}
         \caption{}
         \label{fig:Teaser_pacs}
     \end{subfigure}
     \begin{subfigure}[b]{0.49\linewidth}
         \centering
         \includegraphics[width=\linewidth]{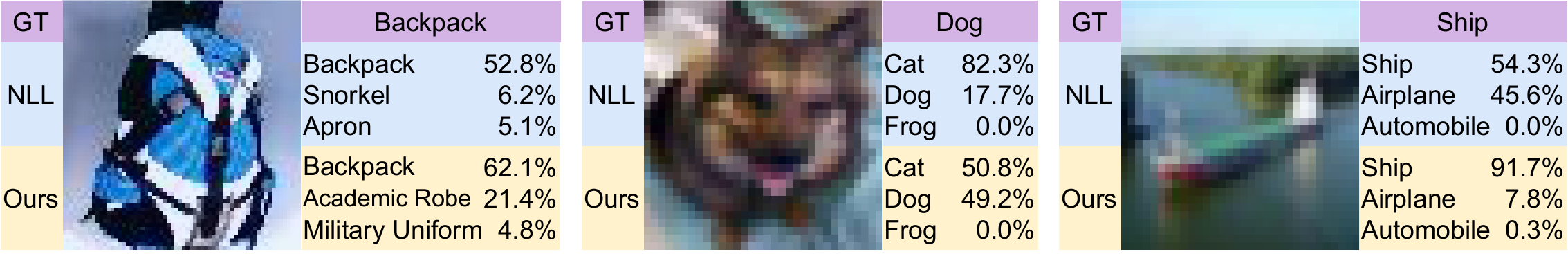}
         \vspace{-0.6cm}
         \caption{}
         \label{fig:Teaser_in_domain}
     \end{subfigure}
     \begin{subfigure}[b]{0.90
     \linewidth}
         \centering
         \includegraphics[width=\linewidth]{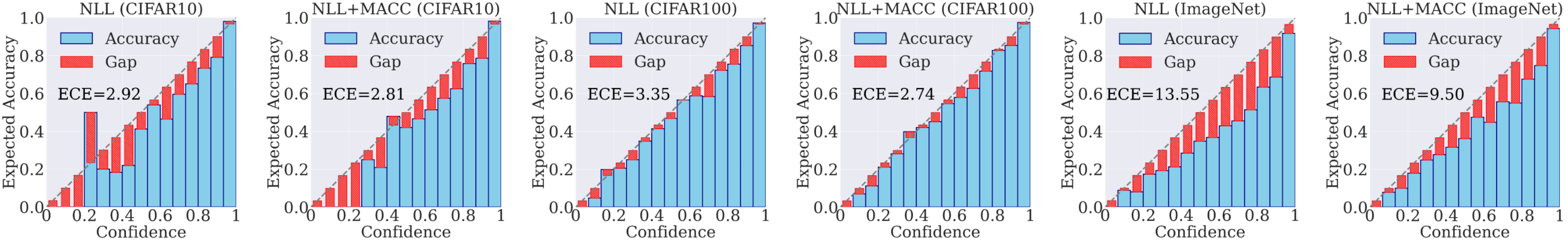}
         \vspace{-0.8cm}
         \caption{}
         \label{fig:Teaser_reliability}
     \end{subfigure}
    \vspace{-0.40cm}
    \caption{We propose a new train-time calibration based on a novel auxiliary loss formulation (MACC). We compare between a model trained with NLL loss and ours (NLL+MACC). (a) shows out-of-domain performance (PACS) while (b) displays in-domain performance (Tiny-ImageNet and CIFAR10). NLL+MACC has higher confidence values for correct predictions (Giraffe (PACS)/Backpack (Tiny-ImageNet)) and lower confidence values for incorrect predictions (Dog (PACS)/Dog (CIFAR10)). (c) Reliability diagrams show that NLL+MACC improves bin-wise miscalibration, thereby alleviating under/over-confident predictions.} 
        
    \label{fig:teaser}
    \vspace{-0.65cm}
\end{figure*}

\section{Introduction}
\label{sec:intro}

Deep neural networks (DNNs) have displayed remarkable performance across many challenging computer vision problems e.g., image classification \cite{dosovitskiy2010image,he2016deep,krizhevsky2017imagenet,simonyan2014very}. However, some recent works \cite{guo2017calibration,mukhoti2020calibrating,ovadia2019can,wenzel2020hyperparameter} have demonstrated that they tend to make over-confident predictions, and so are poorly calibrated. Consequently, the predicted confidences of classes are higher than the actual likelihood of their occurrences. A key reason behind this DNN behaviour is the supervision from zero-entropy signal which trains them to become over-confident. Poorly calibrated models not only create a general suspicion in the model predictions, but more importantly, they can lead to dangerous consequences in many safety-critical applications, including healthcare \cite{dusenberry2020analyzing,sharma2017crowdsourcing}, autonomous vehicles \cite{grigorescu2020survey}, and legal research \cite{yu2019s}. In such applications, providing a correct confidence is as significant as providing a correct label. For instance, in automated healthcare, if the control is not shifted to a doctor when the confidence of the incorrect prediction from a disease diagnosis network is high \cite{jiang2012calibrating}, it can potentially lead to disastrous outcomes.
%
We have seen some recent attempts towards improving the network calibration. Among them, a simple technique is based on a post-hoc procedure, which transforms the outputs of a trained network \cite{guo2017calibration}. The parameters of this transformation are typically learned on a hold-out validation set. Such post-hoc calibration methods are simple and computationally efficient, however, they are architecture and data-dependent \cite{liu2022devil}. Furthermore, in many real-world applications, the availability of a hold-out set is not guaranteed. 
Another route to reducing miscalibration is train-time calibration which tends to involve all model parameters. A dominant approach in train-time calibration methods proposes auxiliary losses that can be added to the task-specific loss (e.g., NLL) to reduce miscalibration \cite{hebbalaguppe2022stitch,liu2022devil,mukhoti2020calibrating,pereyra2017regularizing}. These auxiliary losses aim at either increasing the entropy of the predictive distribution \cite{liu2022devil,mukhoti2020calibrating,pereyra2017regularizing} or aligning the predictive confidence with the predictive accuracy \cite{hebbalaguppe2022stitch,kumar2018trainable}. 

We take the train-time route to calibration, and propose an auxiliary loss function: Multi-class alignment of predictive mean confidence and predictive certainty (MACC). It is founded on the observation that a model's predictive certainty is correlated to its calibration performance. 
So, a higher gap between the predictive mean confidence and predictive certainty translates directly to a greater miscalibration both for in-distribution and out-of-distribution predictions. If a model is confident then it should also produce a relatively low spread in the logit distribution and vice versa. 
Proposed loss function is differentiable, operates on minibatches and is formulated to be used with other task-specific loss functions. Besides showing effectiveness for calibrating in-distribution examples, it is also capable of improving calibration of out-of-distribution examples (Fig.~\ref{fig:teaser}). 
\noindent\textbf{Contributions:}~\textbf{(1)} We empirically observe a correlation between a model's predictive certainty and its calibration performance. ~\textbf{(2)} To this end, we propose a simple, plug-and-play auxiliary loss term (MACC) which attempts to \emph{align the predictive mean confidence with the predictive certainty for all class labels}.~It can be used with other task-specific loss functions, such as Cross Entropy (CE), Label Smoothing (LS) \cite{muller2019does} and Focal Loss (FL) \cite{mukhoti2020calibrating}.~\textbf{(3)} Besides the predicted class label, it also reduces the gap between the certainty and mean confidence for non-predicted class labels, thereby improving the calibration of non-predicted class labels.~\textbf{(4)} We carry out extensive experiments on three in-domain scenarios, CIFAR-10/100\cite{krizhevsky2009learning}, and Tiny-ImangeNet\cite{deng2009imagenet}, a class-imbalanced scenario SVHN \cite{netzer2011reading}, and four out-of-domain scenarios, CIFAR-10/100-C (OOD) \cite{hendrycks2016baseline}, Tiny-ImangeNet-C (OOD) \cite{hendrycks2016baseline} and PACS \cite{li2017deeper}. 
Results show that our loss is consistently more effective than the existing state-of-the-art methods in calibrating both in-domain and out-of-domain predictions. Moreover, we also show the effectiveness of our approach on non-visual pattern recognition task of natural language classification (20 Newsgroups dataset \cite{lang1995newsweeder}) and a medical image classification task (Mendeley dataset \cite{kermany2018labeled}). Finally, we also report results with a vision transformer-based baseline (DeiT-Tiny \cite{touvron2021training}) to show the applicability of our method.





\section{Related Work}
\label{sec:related_work}

\noindent \textbf{Post-hoc calibration methods:} A classic approach for improving model calibration, known as post-hoc calibration, transforms the outputs of a trained model \cite{ding2021local,guo2017calibration,tomani2021post,zhang2020mix}. 
Among different post-hoc calibration methods, a simple technique is temperature scaling (TS) \cite{guo2017calibration}, which is a variant of Platt scaling \cite{guo2017calibration}. It scales the logits (i.e. pre-softmax activations) by a single temperature parameter, which is learned on a hold-out validation set. TS increases the  \mbox{entropy} of the predictive distribution, which is beneficial towards improving model calibration. However, it decreases the confidence of all predictions, including the correct one. TS which relies on a single parameter for transformation can be generalized to a matrix transform, where the matrix is also learnt using a hold-out validation set. Dirichlet calibration (DC) employs Dirichlet distributions for scaling the Beta-calibration \cite{kull2017beta} method to a multi-class setting. 
DC is incorporated as a layer in a neural network on log-transformed class probabilities, which is learnt using a hold-out validation set. Although TS improves model calibration for in-domain predictions, \cite{ovadia2019can} showed that it performs poorly for out-of-domain predictions. To circumvent this, \cite{tomani2021post} proposed to perturb the validation set before performing the post-hoc calibration. 
Recently, \cite{ma2021meta} proposed a ranking model to improve the post-hoc model calibration, and \cite{ding2021local} used a regressor to obtain the temperature parameter at the inference stage.

\noindent \textbf{Train-time calibration methods:} Brier score is considered as one of the earliest train-time calibration technique for binary probabilistic forecast \cite{brier1950verification}. Later, \cite{guo2017calibration} demonstrated that the models trained with negative log-likelihood (NLL) tend to be over-confident, and thus, there is a dissociation between NLL and calibration. Several works proposed auxiliary losses that can be used with NLL to improve miscalibration. For instance, \cite{pereyra2017regularizing} penalized the over-confident predictions by using entropy as a regularization term, and \cite{muller2019does} showed that label smoothing (LS) can improve model calibration. A similar insight was reported by \cite{mukhoti2020calibrating}, that Focal loss (FL) implicitly improves model calibration. It minimizes the KL divergence between the predictive distribution and the target distribution, and at the same time increases the entropy of the predictive distribution. These methods establish that implicit or explicit maximization of entropy improves calibration performance. Based on this observation, \cite{liu2022devil} proposed a calibration technique based on inequality constraints, which imposes a margin between logit distances.
Recently, \cite{liang2020improved} incorporated the difference between confidence and accuracy (DCA) as an auxiliary loss term with the the cross-entropy loss. Similarly, \cite{kumar2018trainable} developed an auxiliary loss term (MMCE), for model calibration that is computed with a reproducing kernel in a Hilbert space \cite{gretton2013introduction}. Prior methods, such as \cite{kumar2018trainable,liang2020improved}, only calibrate the maximum class confidence. To this end, \cite{hebbalaguppe2022stitch} proposed an auxiliary loss term, namely MDCA, that calibrates the non-maximum class confidences along with the maximum class confidence.
We also take the train-time calibration route, however, different to existing methods, we propose to minimize the gap between the predictive mean confidence and predictive certainty to improve model calibration.

\noindent \textbf{Other calibration methods:} Some methods learn to discard OOD samples, either at train-time or post-hoc stage, which reduces over-confidence and leads to improved calibration. Hein et al. \cite{hein2019relu} demonstrated ReLU makes DNNs provide high confidence for an input sample that lies far away from the training samples.
%
%
Guo et al. \cite{guo2017calibration} explored the impact of width, and depth of a DNN, batch normalization, and weight decay on model calibration. 
%
For more literature on calibrating a DNN through OOD detection, we refer the reader to \cite{devries2018learning,hendrycks2018deep,meronen2020stationary,padhy2020revisiting}.

\noindent \textbf{Calibration and uncertainty estimation in DNNs:} 
Many probabilistic approaches emerge from the Bayesian formalism \cite{bernardo2009bayesian}, in which a prior distribution over the neural network (NN) parameters is assumed, and then a training data is used to obtain the posterior distribution over the NN parameters, which is then used to estimate predictive uncertainty. Since the exact Bayesian inference is computationally intractable, several approximate inference techniques have been proposed, including variational inference \cite{blundell2015weight,louizos2016structured}, and stochastic expectation propagation \cite{hernandez2015probabilistic}. 
Ensemble learning is another approach for quantifying uncertainty that uses the empirical variance of the network predictions. We can create ensembles using different techniques. For instance, with the differences in model hyperparameters \cite{wenzel2020hyperparameter}, random initialization of weights and random shuffling of training examples \cite{lakshminarayanan2017simple}, dataset shift \cite{ovadia2019can}, and Monte Carlo (MC) dropout \cite{gal2016dropout,zhang2019confidence}. 
In this work, we chose to use MC dropout \cite{gal2016dropout} to estimate predictive mean confidence and predictive uncertainty of a given example for all class labels. It provides a distribution of class logit scores and is simple to implement. However, the conventional implementation of MC dropout can incur high computational cost for large datasets, architectures, and longer training schedules.
To this end, we resort to an efficient implementation of MC dropout that greatly reduces this computational overhead.  


\section{Proposed Methodology}
\label{sec:method}

\noindent \textbf{Preliminaries:} We consider the task of classification where we have a dataset $\mathcal{D}=\langle(\mathbf{x}_{i},y^{*}_{i})\rangle_{i=1}^{N}$ of $N$ input examples sampled from a joint distribution $\mathcal{D}(\mathcal{X},\mathcal{Y})$, where $\mathcal{X}$ is an input space, and $\mathcal{Y}$ is the label space.
$\mathbf{x}_{i} \in \mathcal{X} \in \mathbb{R}^{H \times W \times C}$ is an input image with height $H$, width $W$, and number of channels $C$. 
Each image has a corresponding ground truth class label $y^{*}_{i} \in \mathcal{Y} = \{1,2,...,K\}$.
Let us denote a classification model $\mathcal{F}_{cls}$, that typically outputs a confidence vector $\mathbf{s}_{i} \in \mathbb{R}^{K}$. Since each element of vector $\mathbf{s}_{i}$ is a valid (categorical) probability, it is considered as the confidence score of the corresponding class label. The predicted class label $\hat{y}_{i}$ can be computed as: $ \hat{y}_{i} = \argmax_{y \in \mathcal{Y}} \mathbf{s}_{i}[y]$.
Likewise, the confidence score of the predicted class $\hat{y}_{i}$ is obtained as: $\hat{s}_{i} = \mathrm{max}_{y \in \mathcal{Y}} \mathbf{s}_{i}[y]$.


\subsection{Definition and Quantification of Calibration}
\noindent\textbf{Definition:} We can define a perfect calibration if the (classification) accuracy for a given confidence score is aligned with this confidence score for all possible confidence scores \cite{guo2017calibration}: $\mathbb{P}(\hat{y} = {y}^{*} | \hat{s} = s) = s \quad \forall s\in[0,1],$
%
where $\mathbb{P}(\hat{y} = y^{*} | \hat{s} = s)$ is the accuracy for a given confidence score $\hat{s}$. The expression only captures the calibration of the predicted label i.e.~associated with the maximum class confidence score $\hat{s}$. The confidence score of non-predicted classes, also called as non-maximum class confidence scores, can also be calibrated. It provides us with a more general definition of perfect calibration and can be expressed as: $ \mathbb{P}(y = y^{*} | \mathbf{s}[y] = s) = s \quad \forall s\in[0,1].$




\noindent \textbf{Expected calibration error (ECE):} ECE is computed by first obtaining the absolute difference between the average confidence of the predicted class and the average accuracy of samples, that are predicted with a particular confidence score. This absolute difference is then converted into a weighted average by scaling it with the relative frequency of samples with a particular confidence score. The above two steps are repeated for all confidence scores and then the resulting weighted averages are summed \cite{naeini2015obtaining}: $\mathrm{ECE} = \sum_{i=1}^{M} \frac{|B_{i}|}{N} \bigg | \frac{1}{|B_{i}|}\sum_{j:\hat{s}_{j}\in B_{i}} \mathbb{I}(\hat{y}_j=y_{j}^{*})-\frac{1}{|B_{i}|}\sum_{j:\hat{s}_{j} \in B_{i}}\hat{s}_{j} \bigg|$.
Where $N$ is the total number of examples. Since the confidence values have a continuous interval, the confidence range $[0,1]$ is divided into $M$ bins. 
$|B_{i}|$ is the number of examples falling in $i^{th}$ confidence bin. $\frac{1}{|B_{i}|}\sum_{j:\hat{s}_{j}\in B_{i}} \mathbb{I}(\hat{y}_j=y_{j}^{*})$ denotes the average accuracy of examples lying in $i^{th}$ bin, and $\frac{1}{|B_{i}|}\sum_{j:\hat{s}_{j} \in B_{i}}\hat{s}_{j}$ represents the average confidence of examples belonging to $i^{th}$ confidence bin. The ECE metric for measuring DNN miscalibration has two limitations. First, the whole confidence vector is not accounted for calibration. Second, due to binning of the confidence interval, the metric is not differentiable. See description on Maximum calibration error (MCE) in supplementary material.


\noindent \textbf{Static calibration error (SCE):} SCE extends ECE by taking into account the whole confidence vector, thereby measuring the calibration performance of non-maximum class confidences \cite{nixon2019measuring}, $\mathrm{SCE} = \frac{1}{K}\sum_{i=1}^{M}\sum_{j=1}^{K} \frac{|B_{i,j}|}{N} \bigg | A_{i,j} - C_{i,j} \bigg|$. Where $K$ represents the number of classes and $|B_{i,j}|$ is the number of examples from the $j^{th}$ class and the $i^{th}$ bin.
$A_{i,j} = \frac{1}{|B_{i,j}|}\sum_{k: \mathbf{s}_{k}[j]\in B_{i,j}} \mathbb{I}(j=y_k)$ denotes the average accuracy and $C_{i,j} = \frac{1}{|B_{i,j}|}\sum_{k: \mathbf{s}_{k}[j]\in B_{i,j}}\mathbf{s}_{k}[j]$ represents the average confidence of the examples belonging to the $j^{th}$ class and the $i^{th}$ bin. Similar to ECE metric, SCE metric is not differentiable, and so it cannot be used as a loss in gradient-based learning methods.



\subsection{Proposed Auxiliary Loss: MACC}
\label{subsec:MACC}


Our auxiliary loss (MACC) aims at reducing the deviation between the predictive mean confidence and the predictive certainty for predicted and non-predicted class labels. 

\noindent \textbf{Quantifying mean confidence and certainty:}
Our proposed loss function requires the estimation of \emph{class-wise mean confidence and certainty}. We choose to use the MC dropout method \cite{gal2016dropout} to estimate both of these quantities because it provides a distribution of (logit) scores for all possible classes and only requires the addition of a single dropout layer $(\mathcal{M})$, which in our case, is added between the feature extractor \textit{f}($\cdot$) that generates features and the classifier \textit{g}($\cdot$) that projects the extracted features into class-wise logits vector.
\begin{figure}[!htp]
    \vspace{-0.5cm}
    \centering
    \includegraphics[width=0.8\linewidth]{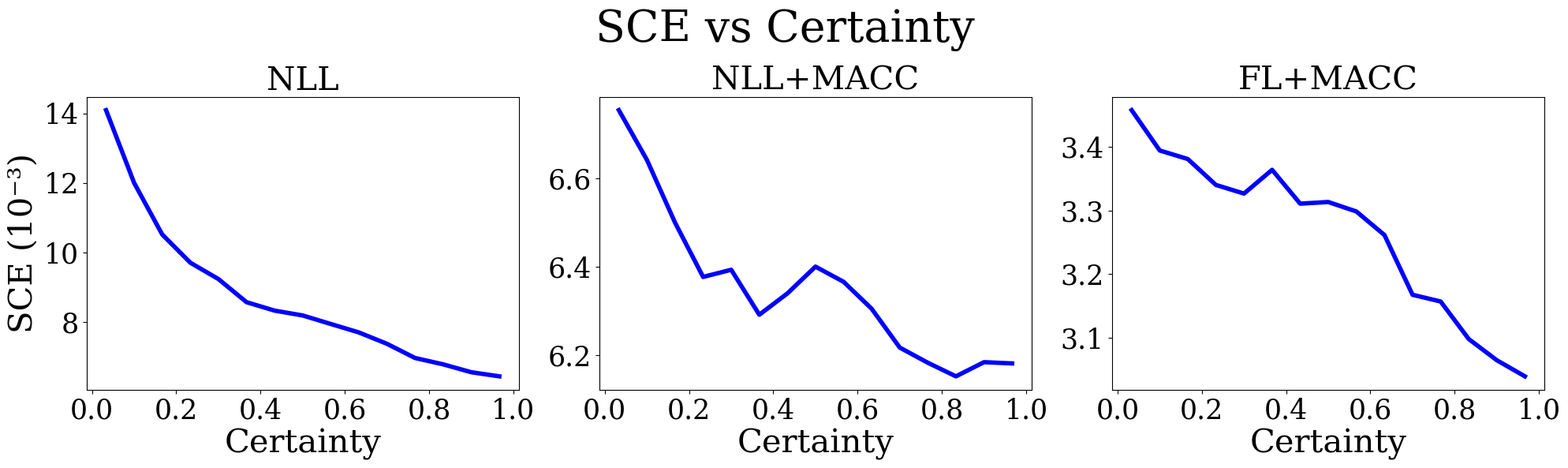}
    \caption{Left:~We investigate if there is a relationship between static calibration error and individual (output) predictive uncertainties. We observe (negative) correlation between a model's predictive certainty and its calibration error (SCE) i.e. as the certainty increases, the calibration error goes down (CIFAR10 trained on ResNet56 model with dropout). Middle \& right: Based on this observation, we propose to align predictive mean confidence with the predictive certainty (NLL/FL+MACC), which allow a rapid reduction in the calibration error in comparison to baseline (NLL). See supplementary material for details of the plot.}
    
     \vspace{-0.70cm}
    \label{fig:motivation_new}
\end{figure}
The conventional implementation of MC dropout technique requires $\mathrm{N}$ MC forward passes for an input example $\mathbf{x}_{i}$ through the model $\mathcal{F}_{cls}$. From the resulting logits distribution, we can then estimate the mean and variance for each class $j$, which reflects for $\mathbf{x}_{i}$, its predictive mean logit score $\bar{\mathbf{z}}_{i}[j]$ and the predictive uncertainty in logit scores $\mathbf{u}_i[j]$, respectively, where $\bar{\mathbf{z}}_{i},\mathbf{u}_{i} \in \mathbb{R}^{K}$. To obtain predictive mean confidence $\bar{\mathbf{s}}_i[j]$, we apply softmax to $\bar{\mathbf{z}}_{i}[j] \forall j$. The certainty $\mathbf{c}_i[j]$ is obtained from the uncertainty $\mathbf{u}_{i}[j]$ as: $\mathbf{c}_i[j] = 1 - \mathrm{tanh}(\mathbf{u}_{i}[j])$. The $\mathrm{tanh}$ is used to scale the uncertainty values between 0 and 1.

We resort to an efficient implementation of MC dropout technique aimed at reducing its computational overhead, which is of concern during model training. Specifically, we feed an input example $\mathbf{x}_{i}$ to the feature extractor network only once and obtain the extracted features $\mathbf{f}_{i}$. These extracted features are then fed to the combination of dropout layer and classifier ($\textit{g}\circ\mathcal{M}$($\mathbf{f}_{i}$)) for $\mathrm{N}$ number of MC forward passes. Specifically, $\mathbf{u}_i[j] = \frac{1}{N-1}\sum_{m=1}^{M}([\textit{g}\circ\mathcal{M}_{m}(\mathbf{f}_{i})]_{j} -  \bar{\mathbf{z}}_{i}[j])^{2}$, where $ \bar{\mathbf{z}}_{i}[j]$ represents the mean of the logit distribution given by: $ \bar{\mathbf{z}}_{i}[j] = \frac{1}{N}\sum_{m=1}^{M}[\textit{g}\circ\mathcal{M}_{m}(\textbf{f}_{i})]_{j}$.
%
This so-called architecture-implicit implementation of MC dropout enjoys the benefit of performing only a single forward pass through the feature extractor $\textit{f}$ as opposed to $\mathrm{N}$ forward passes in the conventional implementation. 
We empirically observe that, for $\mathrm{10}$ MC forward passes, the efficient implementation reduces the overall training time by $\mathrm{7}$ times compared to the conventional implementation (see suppl.). 
Deep ensembles~\cite{lakshminarayanan2017simple} is an alternate to MCDO, however, it is computationally expensive to be used in a train-time calibration approach. On CIFAR10, training deep ensembles with 10 models require around $7.5$ hours whereas ours with 10 forward passes, only require around an hour.

\noindent\textbf{MACC:} The calibration is a frequentist notion of uncertainty and could be construed as a measure reflecting a network's \emph{overall predictive uncertainty}~\cite{lakshminarayanan2017simple}. So, we investigate if there is a relationship between static calibration error and individual (output) predictive uncertainties. We identify a (negative) correlation between a model's predictive certainty and its calibration error (SCE). In other words, as the certainty increases, the calibration error goes down (Fig.~\ref{fig:motivation_new}). With this observation, we propose to align the predictive mean confidence of the model with its predictive certainty. Our loss function is defined as:
\begin{equation}
  \mathcal{L}_{\text{MACC}} = \frac{1}{K}\sum_{j=1}^{K}\left \lvert\frac{1}{M}\sum_{i=1}^{M}\bar{\mathbf{s}}_i[j]-\frac{1}{M}\sum_{i=1}^{M}\mathbf{c}_i[j]\right\rvert,
  \label{eq:mcaa_formulation}
\end{equation}




where $\bar{\mathbf{s}}_i[j]$ denotes the predictive mean confidence of the $i^{th}$ example in the mini-batch belonging to the $j^{th}$ class. Likewise, $\mathbf{c}_i[j]$ represents the certainty of the $i^{th}$ example in the mini-batch belonging to the $j^{th}$ class. $M$ is the number of examples in the mini-batch, and $K$ is the number of classes. 

\begin{figure*}[!htp]
    \vspace{-0.5cm}
     \centering
     \begin{subfigure}[b]{0.24\linewidth}
         \centering
         \includegraphics[width=\linewidth]{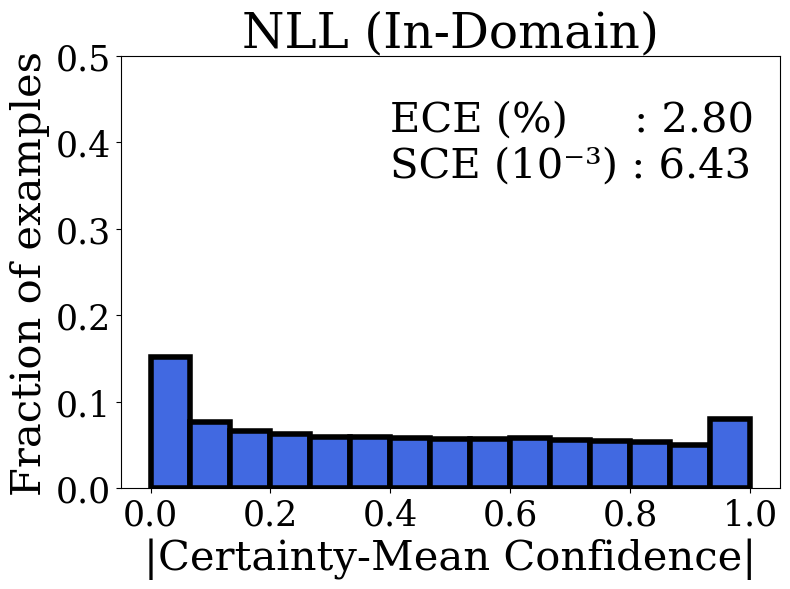}
         \vspace{-0.6cm}
         \caption{}
         \label{fig:nll_id}
     \end{subfigure}
     \begin{subfigure}[b]{0.24\linewidth}
         \centering
         \includegraphics[width=\linewidth]{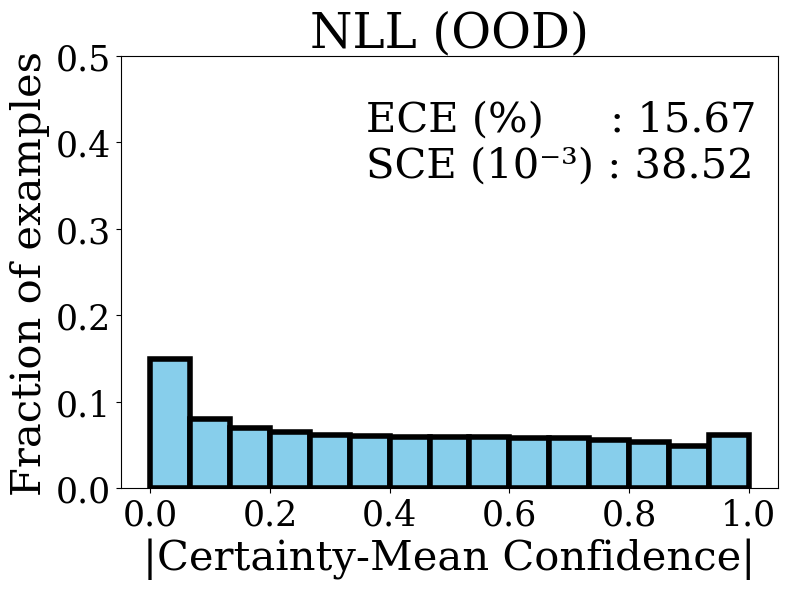}
         \vspace{-0.6cm}
         \caption{}
         \label{fig:nll_ood}
     \end{subfigure}
     \begin{subfigure}[b]{0.24\linewidth}
         \centering
         \includegraphics[width=\linewidth]{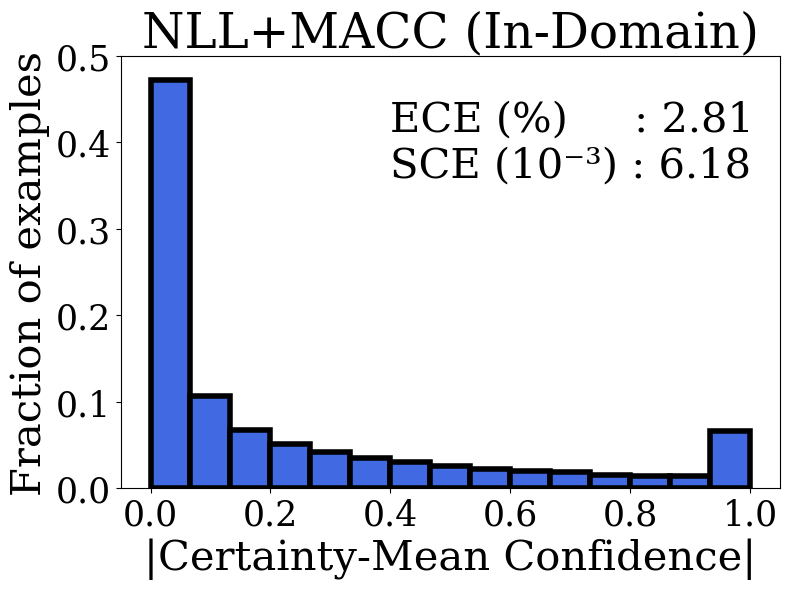}
         \vspace{-0.6cm}
         \caption{}
         \label{fig:nll_mdcu_id}
     \end{subfigure}
     \begin{subfigure}[b]{0.24\linewidth}
         \centering
         \includegraphics[width=\linewidth]{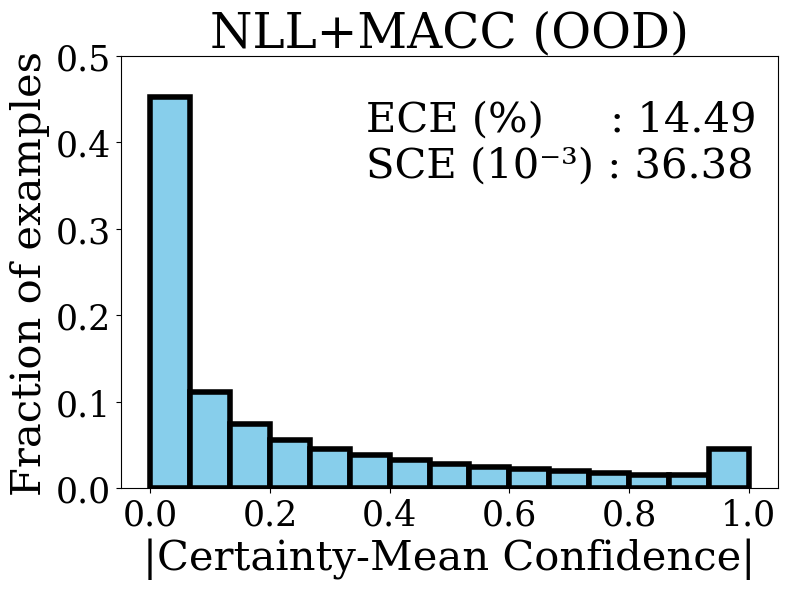}
         \vspace{-0.6cm}
         \caption{}
         \label{fig:nll_mdcu_ood}
     \end{subfigure}
        \vspace{-0.4cm}
        \caption{Empirical distribution of difference between predictive certainty and the predictive mean confidence of all the classes for in-domain examples (CIFAR10), and out-of-domain (OOD) examples (CIFAR10-C). (Left - (a),(b)) When there are relatively greater number of examples with higher gap between the mean confidence and certainty, calibration errors (ECE and SCE) are higher, compared to when there are relatively smaller number of examples (Right - (c),(d)). For (a) and (b), a ResNet56 model with dropout is trained with NLL. For (c) and (d), the same model is trained with NLL+MACC (ours). }
        \vspace{-0.6cm}
        
        
        \label{fig:motivation}
\end{figure*}

\noindent\textbf{Discussion:} Given an example, for which a model predicts high mean confidence, our loss formulation forces the model to also produce relatively low spread in logits distribution and vice versa. This alignment directly helps towards improving the model calibration. Fig.~\ref{fig:motivation_new} shows that, compared to baseline, a model trained with our loss allows a rapid decrease in calibration error. Moreover, Fig.~\ref{fig:nll_id},~\ref{fig:nll_ood} show that when there are relatively greater number of examples with a higher gap between (mean) confidence and certainty (i.e. the distribution is more skewed towards right), 
a model's calibration is poor compared to when there are relatively smaller number of examples with a higher gap (Fig.~\ref{fig:nll_mdcu_id},~~\ref{fig:nll_mdcu_ood}).
The proposed auxiliary loss is a simple, plug-and-play term. It is differentiable and operates over minibatch and thus, it can be used with other task-specific loss functions to improve the model calibration, $\mathcal{L}_{\text{total}} = \mathcal{L}_{\text{task}} + \beta . \mathcal{L}_{\text{MACC}}$,
where $\beta$ represents the weight with which our $\mathcal{L}_{\text{MACC}}$ is added to the task-specific loss function $\mathcal{L}_{\text{task}}$ e.g., CE, LS \cite{muller2019does} and FL \cite{mukhoti2020calibrating}. 

\section{Experiments}
\label{sec:experiments}
\noindent \textbf{Datasets:}
To validate in-domain calibration performance, we use four challenging image classification datasets: CIFAR10\cite{krizhevsky2009learning}, CIFAR100\cite{krizhevsky2009learning}, Tiny-ImageNet\cite{deng2009imagenet} and Mendeley V2\cite{kermany2018labeled} and a natural language processing dataset: 20 Newsgroups\cite{lang1995newsweeder}. Tiny-Image -Net is a subset of ImageNet\cite{imagenet15russakovsky} comprising 200 classes. 
Further, to report calibration performance in out-of-domain scenarios, we show results on four challenging benchmarks: CIFAR10-C\cite{hendrycks2019robustness}, CIFAR100-C\cite{hendrycks2019robustness}, Tiny-ImageNet-C\cite{hendrycks2019robustness} and PACS\cite{li2017deeper}. 
For CIFAR10-C, CIFAR100-C and Tiny-ImageNet-C, we use their corresponding in-domain benchmarks for training and validation. 
Finally, to evaluate calibration performance under class imbalance, we report results on SVHN\cite{netzer2011reading}. 
%

\noindent \textbf{Implementation details and evaluation metrics:}
We use ResNet\cite{he2016deep} and DeiT-Tiny\cite{touvron2021training} (only for CIFAR10) as the backbone networks in our experiments. For our method, we insert a single dropout layer in between the penultimate feature layer and the final classifier of the ResNet architecture. We also input the predictive mean confidence, obtained for our MACC, to the task-specific loss.%
%
We set the number of MC samples to 10 in all experiments. The dropout ratio is sought in the range $p \in \{0.2, 0.3, 0.5\}$ using the validation set. See suppl. material for details. We report the calibration performance with ECE\cite{naeini2015obtaining} and SCE\cite{nixon2019measuring} metrics and the classification performance with top-1 accuracy. The number of bins is  $\mathrm{M = 15}$ for both the calibration metrics across all the experiments. Moreover, we plot reliability diagrams and report AUROC scores. 




\noindent\textbf{Baselines:}
We evaluate MACC against models trained with CE, LS\cite{muller2019does}, FL\cite{lin2017focal}, adaptive sample-dependent focal loss (FLSD)\cite{mukhoti2020calibrating}, brier score (BS)\cite{brier1950verification} and MMCE\cite{kumar2018trainable}. We also compare against the recent auxiliary loss functions: MbLS\cite{liu2022devil} and MDCA\cite{hebbalaguppe2022stitch}. Hyper-parameters of the compared methods are set based on the values reported in the literature. For both MDCA and our loss (MACC), the relative weight is chosen from $\beta \in {\{1,5,10,15,20,25\}}$ and the most accurate model on the validation set is used to report the calibration performance, following MDCA\cite{hebbalaguppe2022stitch} implementation. 
Meanwhile, the scheduled $\gamma$ in FLSD is set to 5 for $s_k \in [0,0.2)$ and 3 for $s_k \in [0.2,1)$, where $s_k$ is the confidence score of the correct class. Refer to the supplementary for the detailed description of these hyperparameters.

\noindent\textbf{Experiments with task-specific loss functions:}
Our loss (MACC) is developed to be used with a task-specific loss function. We consider CE (NLL), LS and FL as the task-specific losses and report the calibration performance with and without incorporating our MACC. For LS we use $\alpha \in \{0.05,0.1\}$ and for FL we use $\gamma \in \{1,2,3\}$ and the most accurate model on the validation set is used to report the performance. 
Table \ref{tab:primary_loss} shows that our auxiliary loss function (MACC) consistently improves the calibration performance of all tasks-specific losses across six datasets. 
We also note that FL is a much stronger task-specific loss function in calibration performance in all datasets, except SVHN and 20 Newsgroups.
The CE loss performs relatively better than FL loss on SVHN and LS performs better on 20 Newsgroups.
We choose to report performance with FL+MACC on all datasets, except SVHN (for which we use CE loss), in all subsequent experiments. 

\noindent\textbf{Comparison with state-of-the-art~(SOTA):}
We compare the calibration performance against recent SOTA train-time calibration methods (Table~\ref{tab:comparison_in_domain},~Table~\ref{tab:rebuttal}). We use NLL+ MbLS to report the performance as it provides better results than FL+MbLS (see suppl.). Our method achieves lower calibration errors in ECE, SCE and AUROC metrics across six datasets. To demonstrate the effectiveness of MACC on natural language classification, we conduct experiments on the 20 Newsgroups dataset (Table~\ref{tab:comparison_in_domain}). Our FL+MACC outperforms others in both SCE and ECE metrics. Experiments with vision-transformer based backbone architecture, namely DeiT-Tiny~\cite{touvron2021training} show that our FL+MACC is capable of improving the calibration performance of DeiT.
Note that, DeIT is a relatively stronger baseline in calibration performance compared to ResNet (see Table~\ref{tab:comparison_in_domain}). For training DeiT models, we use the hyperparameters specified by the authors of DeiT.

\begin{table*}[!htp]
\centering
\tabcolsep=0.07cm
\renewcommand{\arraystretch}{1}
\caption{Calibration performance in SCE ($10^{-3}$) and ECE (\%) metrics of our auxiliary loss (MACC) when added to three task-specific losses: CE, LS, and FL. Throughout, the best results are in bold, and the second best are underlined.} 
\resizebox{0.81\textwidth}{!}{
\begin{tabular}{@{}lccccccccccccc@{}}
\toprule 

\multicolumn{1}{c}{\multirow{2}{*}{\small{\textbf{Dataset}}}} & \multirow{2}{*}{\small{\textbf{Model}}} & \multicolumn{2}{c}{\small{\textbf{NLL}}} & \multicolumn{2}{c}{\small{\textbf{NLL+MACC}}} & \multicolumn{2}{c}{\small{\textbf{LS\cite{muller2019does}}}} & \multicolumn{2}{c}{\small{\textbf{LS+MACC}}} & \multicolumn{2}{c}{\small{\textbf{FL\cite{lin2017focal}}}} & \multicolumn{2}{c}{\small{\textbf{FL+MACC}}} \\ \cline{3-14}

& & {\small{SCE}} & {\small{ECE}} & {\small{SCE}} & {\small{ECE}} & {\small{SCE}} & {\small{ECE}} & {\small{SCE}} & {\small{ECE}} & {\small{SCE}} & {\small{ECE}} & {\small{SCE}} & {\small{ECE}} \\ \midrule

\small{CIFAR10} & \small{ResNet56} & \small{$6.50$} & \small{$2.92$} & \small{$6.18$} & \small{$2.81$} & \small{$5.90$} & \small{$1.85$} & \small{$5.51$} & \small{$1.57$} & \small{$\underline{3.79}$} & \small{$\underline{0.64}$} & \small{$\textbf{3.04}$} & \small{$\textbf{0.59}$} \\
\midrule
\small{CIFAR100} & \small{ResNet56} & \small{$2.01$} & \small{$3.35$} & \small{$\underline{1.99}$} & \small{$2.74$} & \small{$2.08$} & \small{$\underline{0.86}$} & \small{$2.07$} & \small{$0.92$} & \small{$\underline{1.99}$} & \small{$0.89$} & \small{$\textbf{1.97}$} & \small{$\textbf{0.64}$} \\
\midrule
\small{Tiny-ImageNet} & \small{ResNet50} & \small{$2.06$} & \small{$13.55$} & \small{$1.72$} & \small{$9.50$} & \small{$1.50$} & \small{$2.04$} & \small{$\textbf{1.37}$} & \small{$\underline{1.37}$} & \small{$1.50$} & \small{$3.52$} & \small{$\underline{1.44}$} & \small{$\textbf{1.33}$} \\
\midrule
\small{SVHN} & \small{ResNet56} & \small{$\underline{1.70}$} & \small{$\underline{0.43}$} & \small{$\textbf{1.50}$} & \small{$\textbf{0.27}$} & \small{$11.70$} & \small{$4.95$} & \small{$7.72$} & \small{$3.10$} & \small{$7.79$} & \small{$3.55$} & \small{$\underline{1.70}$} & \small{$0.49$} \\
\midrule
\small{20 Newsgroups} & \small{GP CNN} & \small{$23.05$} & \small{$21.11$} & \small{$20.30$} & \small{$18.32$} & \small{$\underline{10.35}$} & \small{$\underline{5.80}$} & \small{$\textbf{9.61}$} & \small{$\textbf{2.17}$} & \small{$21.30$} & \small{${19.54}$} & \small{${13.84}$} & \small{$11.28$} \\
\midrule
\small{Mendeley} & \small{ResNet50} & \small{${206.98}$} & \small{${16.05}$} & \small{${77.12}$} & \small{${7.59}$} & \small{$133.25$} & \small{$5.32$} & \small{$\textbf{57.70}$} & \small{$\underline{4.52}$} & \small{$160.58$} & \small{$5.13$} & \small{$\underline{66.97}$} & \small{$\textbf{4.14}$} \\
\cline{1-14}
\bottomrule
\vspace{-1.2cm}
\end{tabular}
}
\label{tab:primary_loss}

\end{table*}

\begin{table*}[t]
\tabcolsep=0.065cm
\centering
\renewcommand{\arraystretch}{0.90}
\caption{Comparison of calibration performance in SCE ($10^{-3}$) and ECE (\%) metrics with the SOTA train-time calibration methods.}
\resizebox{1\textwidth}{!}{
\begin{tabular}{@{}lccccccccccccccccccc@{}}

\toprule 

\multicolumn{1}{c}{\multirow{2}{*}{\textbf{\small{Dataset}}}} & \multirow{2}{*}{\textbf{\small{Model}}} & \multicolumn{3}{c}{\textbf{\small{BS\cite{brier1950verification}}}} & \multicolumn{3}{c}{\textbf{\small{MMCE\cite{kumar2018trainable}}}} & \multicolumn{3}{c}{\textbf{\small{FLSD\cite{mukhoti2020calibrating}}}} & \multicolumn{3}{c}{\textbf{\small{FL+MDCA\cite{hebbalaguppe2022stitch}}}} &\multicolumn{3}{c}{\textbf{\small{NLL+MbLS\cite{liu2022devil}}}} &\multicolumn{3}{c}{\textbf{\small{FL/NLL+MACC}}}

\\ \cmidrule{3-20}

& & {\small{SCE}} & {\small{ECE}} & {\small{Acc.}} & {\small{SCE}} & {\small{ECE}} & {\small{Acc.}} & {\small{SCE}} & {\small{ECE}} & {\small{Acc.}} & {\small{SCE}} & {\small{ECE}} & {\small{Acc.}} & {\small{SCE}} & {\small{ECE}} & {\small{Acc.}} & {\small{SCE}} & {\small{ECE}} & {\small{Acc.}} \\ \midrule

\multirow{2}{*}{\footnotesize{CIFAR10}} & \footnotesize{ResNet56} & \small{$4.78$} & \small{$1.67$} & \small{$92.46$} & \small{$5.87$} & \small{$1.74$} & \small{$91.75$} & \small{$7.87$} & \small{$3.17$} & \small{$92.37$} & \small{$\underline{3.44}$} & \small{$\underline{0.79}$} & \small{$92.92$} & \small{$4.63$} & \small{$1.48$} & \small{${93.41}$} & \small{$\textbf{3.04}$} & \small{$\textbf{0.59}$} & \small{$92.86$}\\

& \footnotesize{DeiT-Tiny} & \small{$-$} & \small{$-$} & \small{$-$} & \small{$-$} & \small{$-$} & \small{$-$} & \small{$-$} & \small{$-$} & \small{$-$} & \small{$3.63$} & \small{$1.52$} & \small{$97.11$} & \small{$\textbf{3.01}$} & \small{$\underline{0.56}$} & \small{$96.78$} & \small{$\underline{3.02}$} & \small{$\textbf{0.48}$} & \small{$96.43$} \\
\midrule
\footnotesize{CIFAR100} & \footnotesize{ResNet56} & \small{$2.08$} & \small{$4.75$} & \small{$69.64$} & \small{$1.98$} & \small{$2.76$} & \small{$69.71$} & \small{$2.05$} & \small{$1.76$} & \small{${70.97}$} & \small{$\textbf{1.92}$} & \small{$\underline{0.68}$} & \small{$70.34$} & \small{$1.99$} & \small{$0.96$} & \small{${71.33}$} & \small{$\underline{1.97}$} & \small{$\textbf{0.64}$} & \small{$70.50$} \\
\midrule
\footnotesize{Tiny-ImageNet} & \footnotesize{ResNet50} & \small{$-$} & \small{$-$} & \small{$-$} & \small{$-$} & \small{$-$} & \small{$-$} & \small{$1.50$} & \small{$2.75$} & \small{$60.39$} & \small{$\underline{1.44}$} & \small{$2.07$} & \small{$60.24$} & \small{$\textbf{1.42}$} & \small{$\underline{1.59}$} & \small{${62.69}$} & \small{$\underline{1.44}$} & \small{$\textbf{1.33}$} & \small{${61.60}$} \\
\midrule
\footnotesize{SVHN} & \footnotesize{ResNet56} & \small{$2.41$} & \small{$0.51$} & \small{$96.57$} & \small{$12.34$} & \small{$5.88$} & \small{$95.51$} & \small{$17.49$} & \small{$8.59$} & \small{$95.87$} & \small{$1.77$} & \small{$\underline{0.32}$} & \small{$96.10$} & \small{$\textbf{1.43}$} & \small{$0.37$} & \small{${96.59}$} & \small{$\underline{1.50}$} & \small{$\textbf{0.27}$} & \small{${96.74}$} \\
\midrule
\footnotesize{20 Newsgroups} & \footnotesize{GP CNN} & \small{$21.44$} & \small{$18.64$} & \small{$66.08$} & \small{$17.32$} & \small{$14.76$} & \small{$67.54$} & \small{$\underline{14.78}$} & \small{$\underline{11.62}$} & \small{$66.81$} & \small{${17.40}$} & \small{${15.47}$} & \small{$67.04$} & \small{$17.59$} & \small{$15.55$} & \small{$67.74$} & \small{$\textbf{13.84}$} & \small{$\textbf{11.28}$} & \small{$67.87$} \\
\midrule
\footnotesize{Mendeley} & \footnotesize{ResNet50} & \small{$224.34$} & \small{$15.73$} & \small{$76.28$} & \small{$199.16$} & \small{$10.98$} & \small{$78.69$} & \small{$\underline{146.19}$} & \small{$\underline{4.16}$} & \small{$79.17$} & \small{${177.72}$} & \small{${7.85}$} & \small{$78.69$} & \small{${176.93}$} & \small{$9.70$} & \small{$78.85$} & \small{$\textbf{66.97}$} & \small{$\textbf{4.14}$} & \small{$80.93$} \\

\cline{1-20}
\bottomrule
\vspace{-2.5em}
\end{tabular}
}
\label{tab:comparison_in_domain}
\end{table*}

\noindent\textbf{Temperature Scaling (TS):} MACC outperforms NLL/FL + TS (Table~\ref{tab:rebuttal}). We report the best calibration obtained for TS with the primary losses of NLL and FL. For TS we follow the same  protocol used by the MDCA~\cite{hebbalaguppe2022stitch} where 10\% of the training data is set aside as the hold-out validation set and a grid search between the range of 0 to 10 with a step-size of 0.1 is performed to find the optimal temperature value that gives the least NLL on the hold-out set. 
In CIFAR10/100 and SVHN, the obtained metric scores are similar to that of MDCA~\cite{hebbalaguppe2022stitch}. For Tiny-ImageNet, MDCA~\cite{hebbalaguppe2022stitch} does not report results, and our results are better than MbLS~\cite{liu2022devil}, which uses the same protocol as ours.

\begin{table*}[!htp]
\vspace{-0.1cm}
\centering
\tabcolsep=0.08cm
\renewcommand{\arraystretch}{1}
\caption{Calibration performance with Temperature Scaling (TS) \& comparison of calibration performance in AUROC metric with SOTA train-time calibration methods.}
\resizebox{.8\textwidth}{!}{
\begin{tabular}{@{}lccccccccccccc@{}}
\toprule 

\multicolumn{1}{c}{\multirow{3}{*}{\textbf{\footnotesize{Dataset}}}} &
\multicolumn{10}{c}{\textbf{\footnotesize{Comparison with TS}}} & \multicolumn{3}{c}{\textbf{\footnotesize{SOTA Comparison}}}\\ 
& \multicolumn{2}{c}{\textbf{\footnotesize{NLL+MACC}}} & \multicolumn{3}{c}{\textbf{\footnotesize{NLL+TS}}} & \multicolumn{2}{c}{\textbf{\footnotesize{FL+MACC}}} & \multicolumn{3}{c}{\textbf{\footnotesize{FL+TS}}} & \multicolumn{1}{c}{\textbf{\footnotesize{MDCA}}} & \multicolumn{1}{c}{\textbf{\footnotesize{MbLS}}}  & \multicolumn{1}{c}{\textbf{\footnotesize{MACC}}}\\ 
\cmidrule(rl){2-11} \cmidrule(rl){12-14}

& {\footnotesize{SCE}} & {\footnotesize{ECE}} & {\footnotesize{SCE}} & {\footnotesize{ECE}} & {\footnotesize{T}} & {\footnotesize{SCE}} & {\footnotesize{ECE}} & {\footnotesize{SCE}} & {\footnotesize{ECE}} & {\footnotesize{T}} & \multicolumn{3}{c}{\footnotesize{AUROC Score}} \\ 
\cmidrule(rl){1-1}\cmidrule(rl){2-11} \cmidrule(rl){12-14}

\footnotesize{CIFAR10} 
& \footnotesize{$6.18$} & \footnotesize{$2.81$} & \footnotesize{${4.12}$} & \footnotesize{${0.87}$} & \footnotesize{$1.4$} & \footnotesize{$3.04$} & \footnotesize{$0.59$} & \footnotesize{${3.79}$} & \footnotesize{${0.64}$} & \footnotesize{$1.0$} & \footnotesize{$\textbf{0.9966}$} & \footnotesize{${0.9958}$} & \footnotesize{$\textbf{0.9966}$} \\
\footnotesize{CIFAR100} 
& \footnotesize{$1.99$} & \footnotesize{$2.74$} & \footnotesize{${1.84}$} & \footnotesize{${1.36}$} & \footnotesize{$1.1$} & \footnotesize{$1.97$} & \footnotesize{${0.64}$} & \footnotesize{${1.99}$} & \footnotesize{$0.89$} & \footnotesize{$1.0$} & \footnotesize{$\textbf{0.9922}$} & \footnotesize{${0.9916}$} & \footnotesize{$\textbf{0.9922}$}\\
\footnotesize{Tiny-ImageNet} 
& \footnotesize{$1.72$} & \footnotesize{$9.50$} & \footnotesize{${2.06}$} & \footnotesize{${13.55}$} & \footnotesize{$1.0$} & \footnotesize{$1.44$} & \footnotesize{${1.33}$} & \footnotesize{${2.42}$} & \footnotesize{$18.05$} & \footnotesize{$0.6$} & \footnotesize{${0.9848}$} & \footnotesize{${0.9811}$} & \footnotesize{$\textbf{0.9858}$}\\
\footnotesize{SVHN} 
& \footnotesize{$1.50$} & \footnotesize{$0.27$} & \footnotesize{${2.80}$} & \footnotesize{${1.01}$} & \footnotesize{$1.2$} & \footnotesize{$1.70$} & \footnotesize{${49}$} & \footnotesize{${3.00}$} & \footnotesize{$0.91$} & \footnotesize{$0.8$} & \footnotesize{${0.9973}$} & \footnotesize{$\textbf{0.9977}$} & \footnotesize{$\textbf{0.9977}$}\\
\cline{1-14}
\bottomrule
\end{tabular}
}
\vspace{-0.1cm}
\label{tab:rebuttal}
\end{table*}

\noindent\textbf{Class-wise calibration performance and test accuracy:}
Table~\ref{tab:classwise_SVHN} reports class-wise ECE (\%) scores of competing calibration approaches, including MDCA and MbLS, on SVHN and CIFAR10 datasets, with ResNet56. 
In SVHN, NLL+MACC (ours) achieves the lowest ECE(\%) in three classes while demonstrating the second best score in other four. 
FL+ MACC provides the best values in two classes and the second best values in another two classes. 
NLL+MbLS also performs well, being the best in five classes.
In CIFAR10, FL+MACC (ours) provides the best ECE(\%) scores in five classes while showing the second best in all others. Table~\ref{tab:comparison_in_domain} shows the discriminative performance (top-1 accuracy \%) of our loss (MACC) along with the other competing approaches. Our loss shows superior accuracy than most of the existing losses, including MDCA, in CIFAR100 and Tiny-ImageNet. Moreover, it provides the best accuracy in SVHN, Mendeley and 20 Newsgroups.


\begin{table*}[!htp]
\vspace{-0.7cm}
\footnotesize
\tabcolsep=0.05cm
\centering
\renewcommand{\arraystretch}{.99}
\caption{Class-wise calibration performance in ECE($\%$) of competing approaches on SVHN and CIFAR10 benchmarks (ResNet56).}
\resizebox{.99\textwidth}{!}{
\begin{tabular}{@{}lccccccccccccccccccccc@{}}

\toprule 

\multicolumn{1}{c}{\multirow{2}{*}{\textbf{\small{Loss}}}} & \multicolumn{20}{c}{\textbf{\small{Classes}}}
\\ \cline{2-21}
& {\small{0}} & {\small{1}} & {\small{2}} &  {\small{3}} & {\small{4}} & {\small{5}} &  {\small{6}} & {\small{7}} & {\small{8}} & {\small{9}} & {\small{0}} & {\small{1}} & {\small{2}} &  {\small{3}} & {\small{4}} & {\small{5}} &  {\small{6}} & {\small{7}} & {\small{8}} & {\small{9}}
\\ \cline{1-21}
& \multicolumn{10}{c}{\textbf{\small{SVHN}}} & \multicolumn{10}{c}{\textbf{\small{CIFAR10}}}
\\ \cline{2-21}
\textbf{\footnotesize{NLL+MDCA}} & \footnotesize{$0.17$} & \footnotesize{$0.20$} & \footnotesize{$0.34$} & \footnotesize{$0.22$} & \footnotesize{$\underline{0.13}$} & \footnotesize{$\underline{0.16}$} & \footnotesize{$0.15$} & \footnotesize{$0.17$} & \footnotesize{$\underline{0.16}$} & \footnotesize{$\underline{0.14}$}\\
\textbf{\footnotesize{NLL+MACC}} & \footnotesize{$\underline{0.12}$} & \footnotesize{$\underline{0.16}$} & \footnotesize{$\textbf{0.14}$} & \footnotesize{$\underline{0.20}$} & \footnotesize{$0.14$} & \footnotesize{$\textbf{0.14}$} & \footnotesize{$\textbf{0.10}$} & \footnotesize{$0.16$} & \footnotesize{$0.19$} & \footnotesize{$\underline{0.14}$}\\
\textbf{\footnotesize{FL+MDCA}} & \footnotesize{$0.13$} & \footnotesize{$0.22$} & \footnotesize{$0.21$} & \footnotesize{$\underline{0.20}$} & \footnotesize{$0.16$} & \footnotesize{$0.18$} & \footnotesize{$0.16$} & \footnotesize{$\underline{0.14}$} & \footnotesize{$\underline{0.16}$} & \footnotesize{$0.22$} & \footnotesize{$\underline{0.33}$} & \footnotesize{$\textbf{0.32}$} & \footnotesize{$0.42$} & \footnotesize{$0.72$} & \footnotesize{$\textbf{0.21}$} & \footnotesize{$\underline{0.39}$} & \footnotesize{$\underline{0.30}$} & \footnotesize{$\textbf{0.21}$} & \footnotesize{$\textbf{0.21}$} & \footnotesize{$\underline{0.33}$}\\
\textbf{\footnotesize{NLL+MbLS}} & \footnotesize{$\textbf{0.07}$} & \footnotesize{$\textbf{0.14}$} & \footnotesize{$\underline{0.18}$} & \footnotesize{$0.22$} & \footnotesize{$\textbf{0.11}$} & \footnotesize{$0.17$} & \footnotesize{$\underline{0.13}$} & \footnotesize{$\textbf{0.09}$} & \footnotesize{$0.18$} & \footnotesize{$\textbf{0.13}$} & \footnotesize{$\textbf{0.24}$} & \footnotesize{$0.51$} & \footnotesize{$\underline{0.33}$} & \footnotesize{$\underline{0.68}$} & \footnotesize{$0.44$} & \footnotesize{$0.51$} & \footnotesize{$0.48$} & \footnotesize{$0.57$} & \footnotesize{$0.46$} & \footnotesize{$0.41$}\\
\textbf{\footnotesize{FL+MACC}} & \footnotesize{$0.17$} & \footnotesize{$0.23$} & \footnotesize{$\underline{0.18}$} & \footnotesize{$\textbf{0.18}$} & \footnotesize{$0.17$} & \footnotesize{$0.19$} & \footnotesize{$0.17$} & \footnotesize{$\underline{0.14}$} & \footnotesize{$\textbf{0.10}$} & \footnotesize{$0.17$} & \footnotesize{$\underline{0.33}$} & \footnotesize{$\underline{0.34}$} & \footnotesize{$\textbf{0.31}$} & \footnotesize{$\textbf{0.45}$} & \footnotesize{$\underline{0.30}$} & \footnotesize{$\textbf{0.33}$} & \footnotesize{$\textbf{0.25}$} & \footnotesize{$\underline{0.28}$} & \footnotesize{$\underline{0.23}$} & \footnotesize{$\textbf{0.23}$}\\
\cline{1-21}

\bottomrule
\vspace{-0.5cm}
\end{tabular}
}
\label{tab:classwise_SVHN}
\end{table*}

\noindent\textbf{Out-of-distribution performance:}
Table~\ref{tab:comparison_out_domain_cifar} reports the out-of-domain calibration performance on the CIFAR10-C, CIFAR100-C and Tiny-ImageNet-C benchmarks. In both CIFAR10-C and CIFAR100-C datasets, our loss records the best calibration performance in ECE and SCE metrics. In Tiny-ImageNet-C, our loss shows the lowest ECE score and reveal the second lowest SCE score.
We plot calibration performance as a function of corruption level in CIFAR10-C dataset (see Fig.~4 suppl.). Our loss consistently obtains lowest ECE and SCE across all corruption levels.
For the OOD evaluation, including CIFAR10-C, CIFAR100- C, and Tiny-ImageNet-C, we follow the same protocol and train/val splits as used for in-domain evaluation. Specifically, we train a model using the training split, and optimize parameters using the validation split and the trained model is then evaluated on the in-domain test set or the corrupted test set. 




\begin{table*}[!htp]
\vspace{-0.5cm}
\tabcolsep=0.06cm
\centering
\renewcommand{\arraystretch}{1}
\caption{Out of Domain (OOD) calibration performance of competing approaches across CIFAR10-C, CIFAR100-C and Tiny-ImageNet-C.}
\resizebox{0.99\textwidth}{!}{
\begin{tabular}{@{}lcccccccccc@{}}

\toprule 

\multicolumn{1}{c}{\multirow{2}{*}{\textbf{\small{Dataset}}}} & \multirow{2}{*}{\textbf{\small{Model}}} &  \multicolumn{3}{c}{\textbf{\small{FL+MDCA}}} &\multicolumn{3}{c}{\textbf{\small{NLL+MbLS}}} &\multicolumn{3}{c}{\textbf{\small{FL+MACC}}}

\\ \cline{3-11}

& & {\small{SCE ($10^{-3}$)}} & {\small{ECE ($\%$)}} & {\small{Acc. ($\%$)}} & {\small{SCE ($10^{-3}$)}} & {\small{ECE ($\%$)}} & {\small{Acc. ($\%$)}} & {\small{SCE ($10^{-3}$)}} & {\small{ECE ($\%$)}} & {\small{Acc. ($\%$)}}  \\ \midrule

\footnotesize{CIFAR10 (In-Domain)} & \multirow{2}{*}{\small{ResNet56}} & \small{$\underline{3.44}$} & \small{$\underline{0.79}$} & \small{${92.92}$} & \small{$4.63$} & \small{$1.48$} & \small{${93.41}$} & \small{$\textbf{3.04}$} & \small{$\textbf{0.59}$} & \small{$92.86$} \\
\footnotesize{CIFAR10-C (OOD)} & & \small{$29.01$} & \small{$\underline{11.51}$} & \small{$71.30$} & \small{$\underline{27.61}$} & \small{$12.21$} & \small{${73.75}$} & \small{$\textbf{23.71}$} & \small{$\textbf{9.10}$} & \small{${72.85}$} \\
\cline{1-11}
\footnotesize{CIFAR100 (In-Domain)} & \multirow{2}{*}{\small{ResNet56}} & \small{$\textbf{1.92}$} & \small{$\underline{0.68}$} & \small{$70.34$} & \small{$1.99$} & \small{$0.96$} & \small{${71.33}$} & \small{$\underline{1.97}$} & \small{$\textbf{0.64}$} & \small{${70.5}$} \\
\footnotesize{CIFAR100-C (OOD)} & & \small{$4.09$} & \small{$\underline{12.21}$} & \small{$44.74$} & \small{$\underline{4.03}$} & \small{$12.48$} & \small{${45.60}$} & \small{$\textbf{4.01}$} & \small{$\textbf{12.11}$} & \small{${44.90}$} \\
\cline{1-11}
\footnotesize{Tiny-ImageNet (In-Domain)} & \multirow{2}{*}{\small{ResNet50}} & \small{$1.44$} & \small{$2.07$} & \small{$60.24$} & \small{$\textbf{1.42}$} & \small{$\underline{1.59}$} & \small{${62.69}$} & \small{$\underline{1.44}$} & \small{$\textbf{1.33}$} & \small{${61.60}$} \\
\footnotesize{Tiny-ImageNet-C (OOD)} & & \small{$3.87$} & \small{$22.79$} & \small{$20.74$} & \small{$\textbf{3.04}$} & \small{$\underline{18.17}$} & \small{${23.70}$} & \small{$\underline{3.46}$} & \small{$\textbf{17.82}$} & \small{${21.29}$} \\
\cline{1-11}

\bottomrule
\end{tabular}
}
\vspace{-0.7cm}
\label{tab:comparison_out_domain_cifar}
\end{table*}

%
%
We also show the OOD calibration performance on the PACS dataset under two different evaluation protocols. In first, following \cite{hebbalaguppe2022stitch}, a model is trained on \textbf{Photo} domain while \textbf{Art} domain is used as the validation set, and the trained model is then tested on the rest of domains. Table~\ref{tab:comparison_out_domain_pacs} shows that the proposed loss obtains the best calibration performance in ECE score, while the second best in SCE. 
In second, a model is trained on each domain and then tested on all other  domains. In this protocol, $20\%$ of images corresponding to the training domain is randomly sampled as the validation set, and the remaining $3$ domains form the test set.
Table~\ref{tab:comparison_out_domain_pacs_one_v_all} shows that FL+MACC provides improved calibration than all other competing approaches.

\begin{table*}[!htp]
\vspace{-0.4cm}
\small
\tabcolsep=0.07cm
\centering
\caption{OOD calibration performance (SCE ($10^{-2}$) \& ECE (\%)) on PACS when ResNet18 model is trained on \textbf{Photo}, validated on \textbf{Art}, and tested on \textbf{Sketch} and \textbf{Cartoon} \cite{hebbalaguppe2022stitch}.}
\resizebox{0.65\textwidth}{!}{
\begin{tabular}{@{}lccccccccc@{}}
\toprule 
\multirow{2}{*}{\textbf{\small{Domain}}} & \multicolumn{3}{c}{\textbf{\small{FL+MDCA}}} &  \multicolumn{3}{c}{\textbf{\small{NLL+MbLS}}} & \multicolumn{3}{c}{\textbf{\small{FL+MACC}}}
\\ \cline{2-10}
& {\small{Acc.}} & {\small{SCE}} & {\small{ECE}} &  {\small{Acc.}} & {\small{SCE}} & {\small{ECE}} &  {\small{Acc.}} & {\small{SCE}} & {\small{ECE}}
\\ \midrule

\textbf{\small{Cartoon}} & \footnotesize{$25.34$} & \footnotesize{$15.62$} & \footnotesize{$44.05$} & \footnotesize{${27.69}$} & \footnotesize{$\underline{12.82}$} & \footnotesize{$\underline{30.88}$} & \footnotesize{${32.17}$} & \footnotesize{$\textbf{10.19}$} & \footnotesize{$\textbf{22.18}$}\\
\textbf{\small{Sketch}} & \footnotesize{${29.14}$} & \footnotesize{$14.82$} & \footnotesize{$40.16$} & \footnotesize{${26.57}$} & \footnotesize{$\textbf{11.40}$} & \footnotesize{$\textbf{18.03}$} & \footnotesize{$24.94$} & \footnotesize{$\underline{12.59}$} & \footnotesize{$\underline{28.14}$} \\
\cline{1-10}
\textbf{\small{Average}} & \footnotesize{${27.24}$} & \footnotesize{$15.22$} & \footnotesize{$42.10$} & \footnotesize{$27.13$} & \footnotesize{$\underline{12.11}$} & \footnotesize{$\textbf{24.45}$} & \footnotesize{${28.55}$} & \footnotesize{$\textbf{11.39}$} & \footnotesize{$\underline{25.16}$} \\
\cline{1-10}

\bottomrule
\vspace{-0.5cm}
\end{tabular}
}
\label{tab:comparison_out_domain_pacs}
\end{table*}

\begin{table}[!htp]
\vspace{-0.4cm}
\tabcolsep=0.10cm
\centering
\caption{Out of Domain (OOD) calibration performance (SCE ($10^{-2}$) \& ECE (\%)) on PACS when ResNet18 model is trained on each domain and tested on other $3$ domains. Validation set comprises $20\%$ randomly sampled images from the training domain.}
\renewcommand{\arraystretch}{0.95}
\resizebox{0.99\textwidth}{!}{
\begin{tabular}{@{}lccccccccccccccccc@{}}

\toprule 

\multicolumn{3}{c}{\textbf{\small{FL+MDCA}}} &  \multicolumn{3}{c}{\textbf{\small{NLL+MbLS}}} & \multicolumn{3}{c}{\textbf{\small{FL+MACC}}} & \multicolumn{3}{c}{\textbf{\small{FL+MDCA}}} &  \multicolumn{3}{c}{\textbf{\small{NLL+MbLS}}} & \multicolumn{3}{c}{\textbf{\small{FL+MACC}}}
\\ \toprule 
{\small{Acc.}} & {\small{SCE}} & {\small{ECE}} &  {\small{Acc.}} & {\small{SCE}} & {\small{ECE}} &  {\small{Acc.}} & {\small{SCE}} & {\small{ECE}} & {\small{Acc.}} & {\small{SCE}} & {\small{ECE}} &  {\small{Acc.}} & {\small{SCE}} & {\small{ECE}} &  {\small{Acc.}} & {\small{SCE}} & {\small{ECE}}
\\ \toprule 

\multicolumn{9}{c}{\textbf{\small{Photo Domain}}} & \multicolumn{9}{c}{\textbf{\small{Art Domain}}}\\
\footnotesize{${38.48}$} & \footnotesize{$13.59$} & \footnotesize{$38.93$} & \footnotesize{${35.32}$} & \footnotesize{$\underline{13.44}$} & \footnotesize{$\underline{33.77}$} & \footnotesize{$34.75$} & \footnotesize{$\textbf{9.09}$} & \footnotesize{$\textbf{14.63}$}

& \footnotesize{${56.87}$} & \footnotesize{$\underline{9.33}$} & \footnotesize{$24.81$} & \footnotesize{${57.46}$} & \footnotesize{$\textbf{8.06}$} & \footnotesize{$\textbf{15.21}$} & \footnotesize{$51.45$} & \footnotesize{$9.70$} & \footnotesize{$\underline{16.28}$}\\
\cline{1-18}
\cline{1-18}

\multicolumn{9}{c}{\textbf{\small{Cartoon Domain}}} & \multicolumn{9}{c}{\textbf{\small{Sketch Domain}}}\\
\footnotesize{${62.68}$} & \footnotesize{$\textbf{5.79}$} & \footnotesize{$13.91$} & \footnotesize{${59.43}$} & \footnotesize{$\underline{6.38}$} & \footnotesize{$\underline{9.62}$} & \footnotesize{$57.47$} & \footnotesize{$6.58$} & \footnotesize{$\textbf{5.95}$}

& \footnotesize{${18.31}$} & \footnotesize{$\underline{17.99}$} & \footnotesize{$\underline{54.37}$} & \footnotesize{$11.57$} & \footnotesize{$19.50$} & \footnotesize{$57.17$} & \footnotesize{${13.24}$} & \footnotesize{$\textbf{16.68}$} & \footnotesize{$\textbf{45.23}$}\\
\cline{1-18}

& & & & & \multicolumn{9}{c}{\textbf{\small{Average}}}\\
& & & & & \footnotesize{${44.09}$} & \footnotesize{$\underline{11.68}$} & \footnotesize{$33.00$} & \footnotesize{${40.94}$} & \footnotesize{$11.83$} & \footnotesize{$\underline{28.94}$} & \footnotesize{$39.23$} & \footnotesize{$\textbf{10.51}$} & \footnotesize{$\textbf{20.52}$}\\
\cline{6-14}

\bottomrule
\vspace{-1.2cm}
\end{tabular}
}
\label{tab:comparison_out_domain_pacs_one_v_all}
\end{table}

\begin{figure}[!htp]
     \centering
     \vspace{-0.5cm}
     \begin{subfigure}[b]{0.24\linewidth}
         \centering
\includegraphics[width=\linewidth]{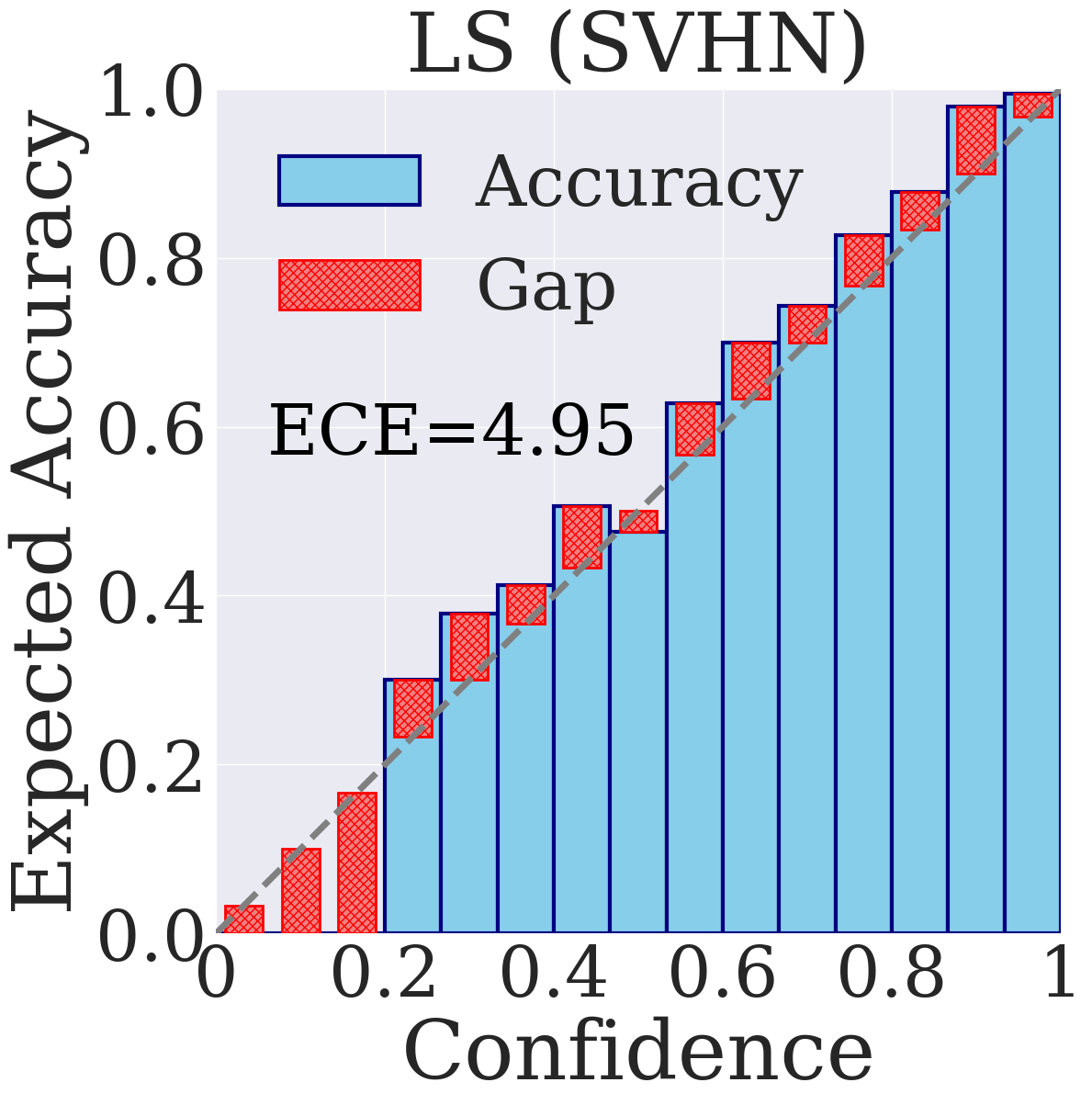}
         \vspace{-0.6cm}
         \caption{}
         \label{fig:under_conf_reliability}
     \end{subfigure}
     \begin{subfigure}[b]{0.24\linewidth}
         \centering
         \includegraphics[width=\linewidth]{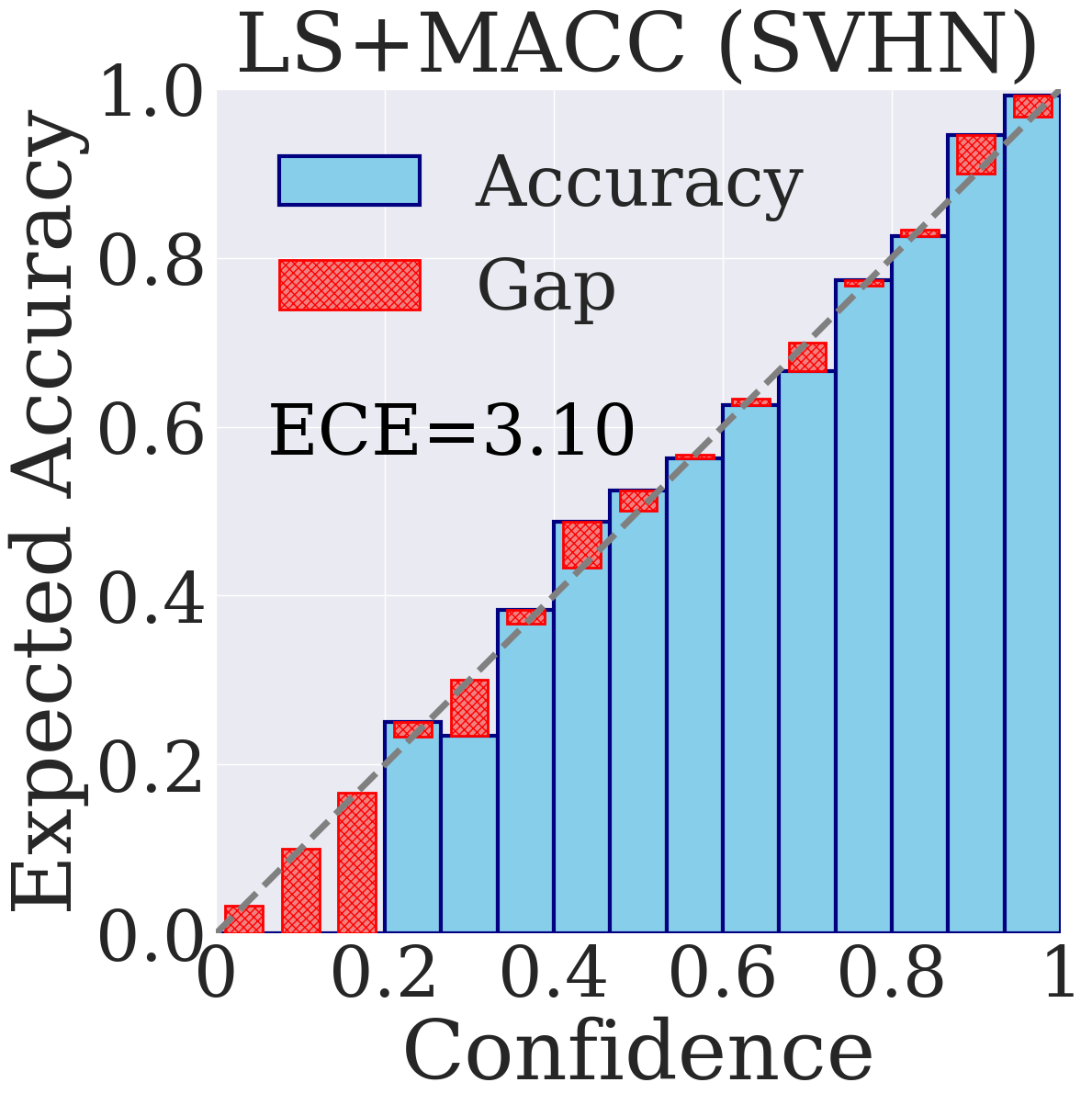}
         \vspace{-0.6cm}
         \caption{}
         \label{fig:under_conf_corrected_reliability}
     \end{subfigure}
     \begin{subfigure}[b]{0.24\linewidth}
         \centering
         \includegraphics[width=\linewidth]{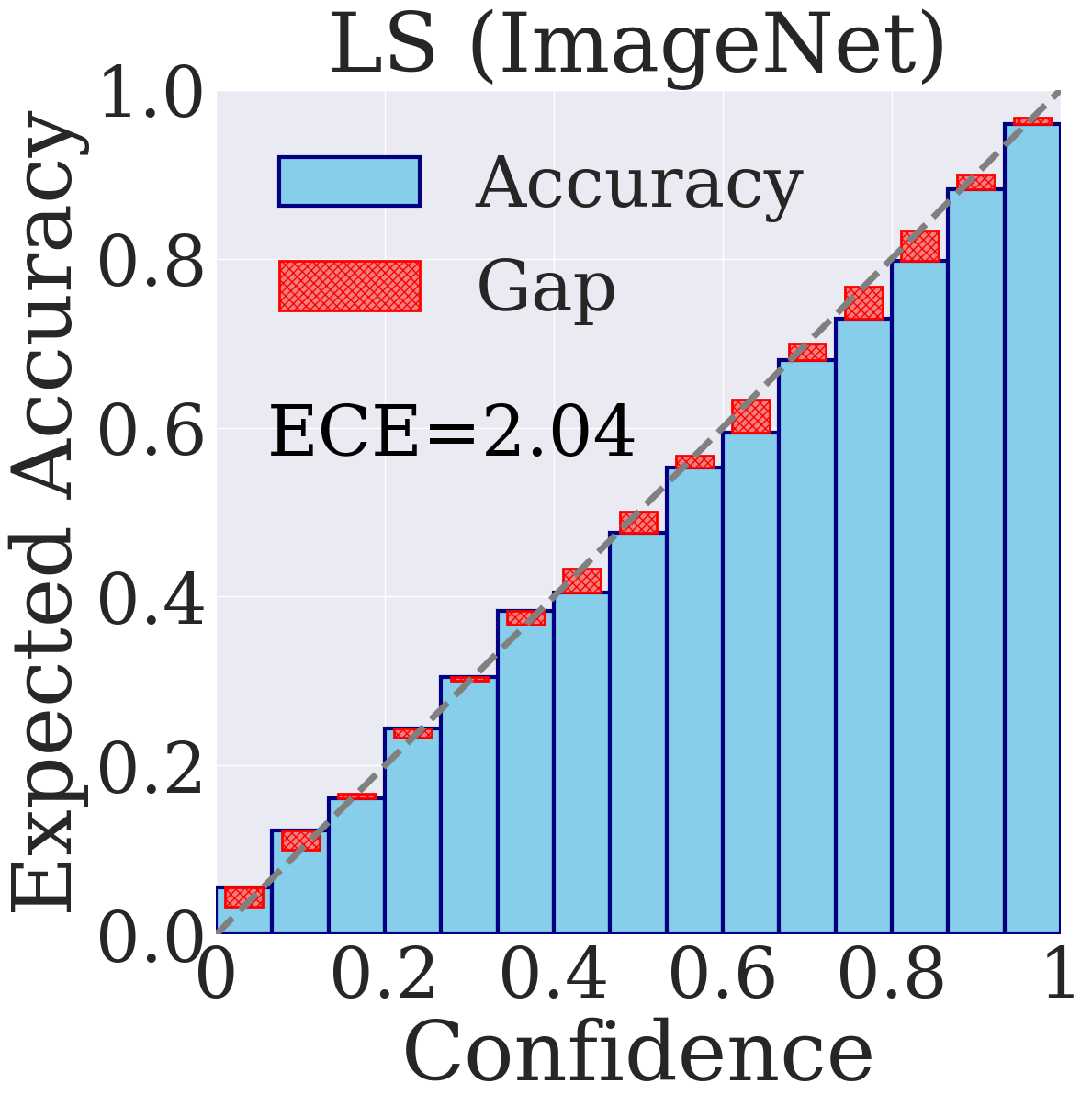}
         \vspace{-0.6cm}
         \caption{}
         \label{fig:over_conf_reliability}
     \end{subfigure}
     \begin{subfigure}[b]{0.24\linewidth}
         \centering
         \includegraphics[width=\linewidth]{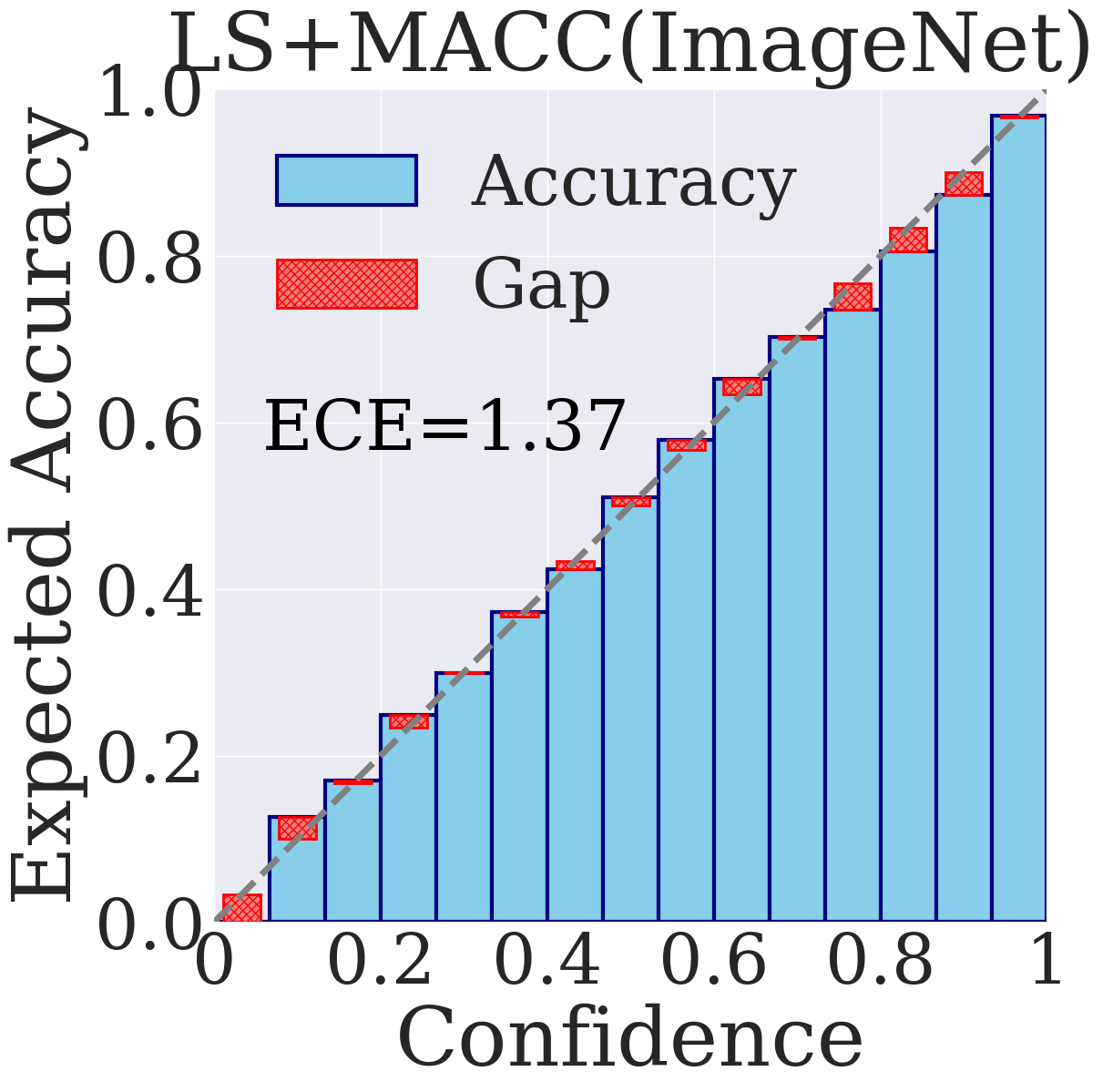}
         \vspace{-0.6cm}
         \caption{}
         \label{fig:over_conf_corrected_reliability}
     \end{subfigure}
     \begin{subfigure}[b]{0.24\linewidth}
         \centering
         \includegraphics[width=\linewidth]{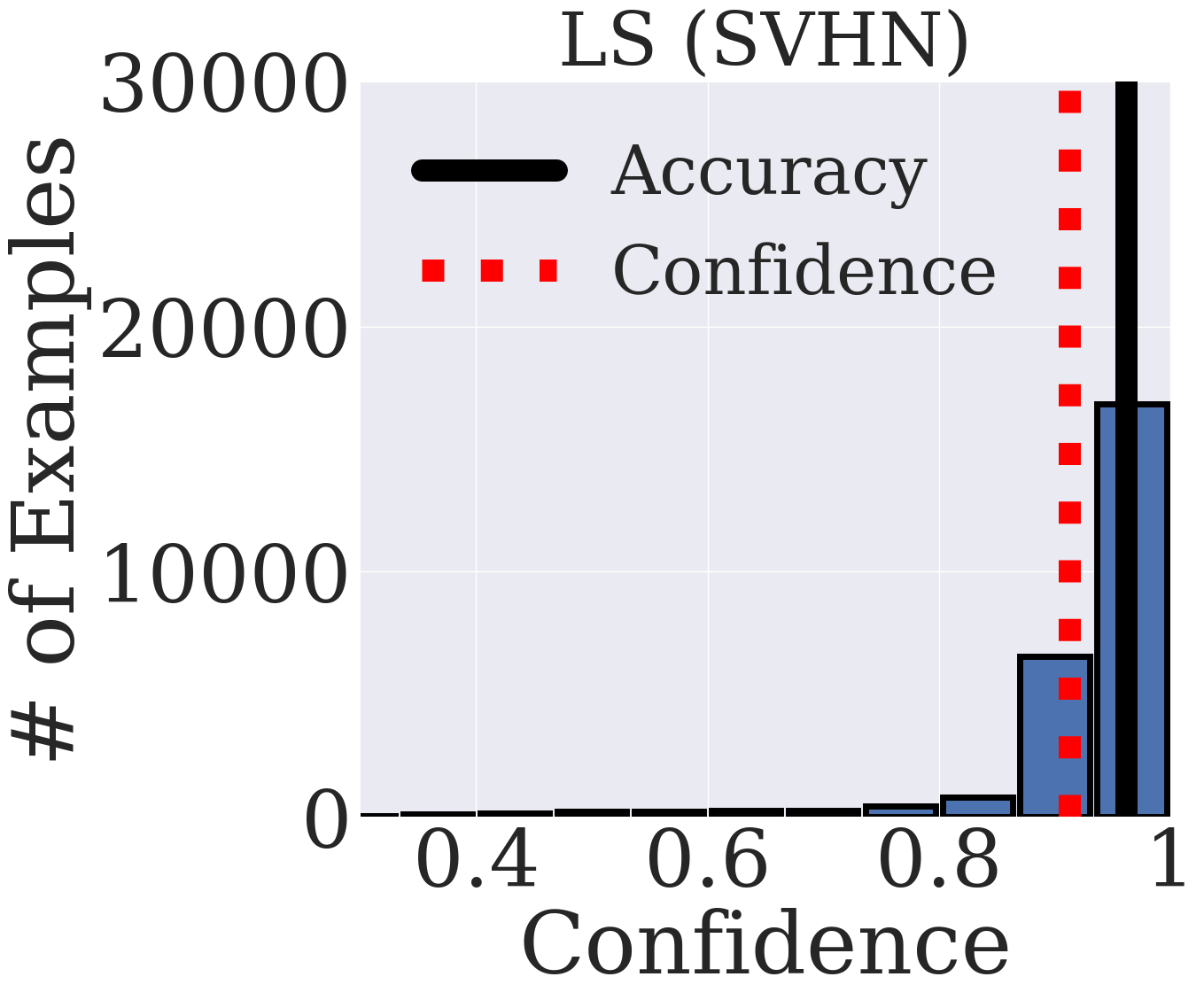}
         \vspace{-0.6cm}
         \caption{}
         \label{fig:under_conf}
     \end{subfigure}
     \begin{subfigure}[b]{0.24\linewidth}
         \centering
         \includegraphics[width=\linewidth]{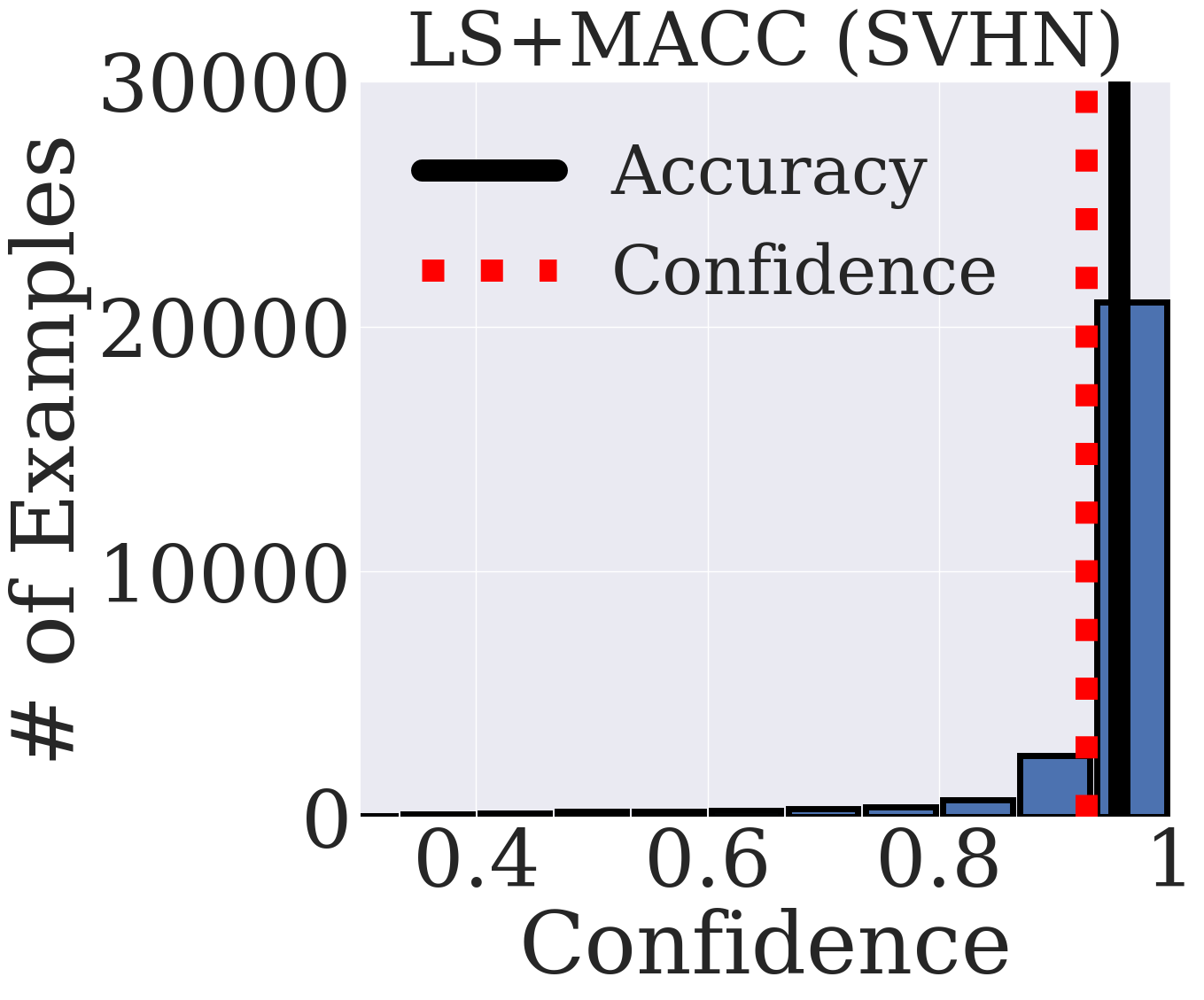}
         \vspace{-0.6cm}
         \caption{}
         \label{fig:under_conf_corrected}
     \end{subfigure}
     \begin{subfigure}[b]{0.24\linewidth}
         \centering
         \includegraphics[width=\linewidth]{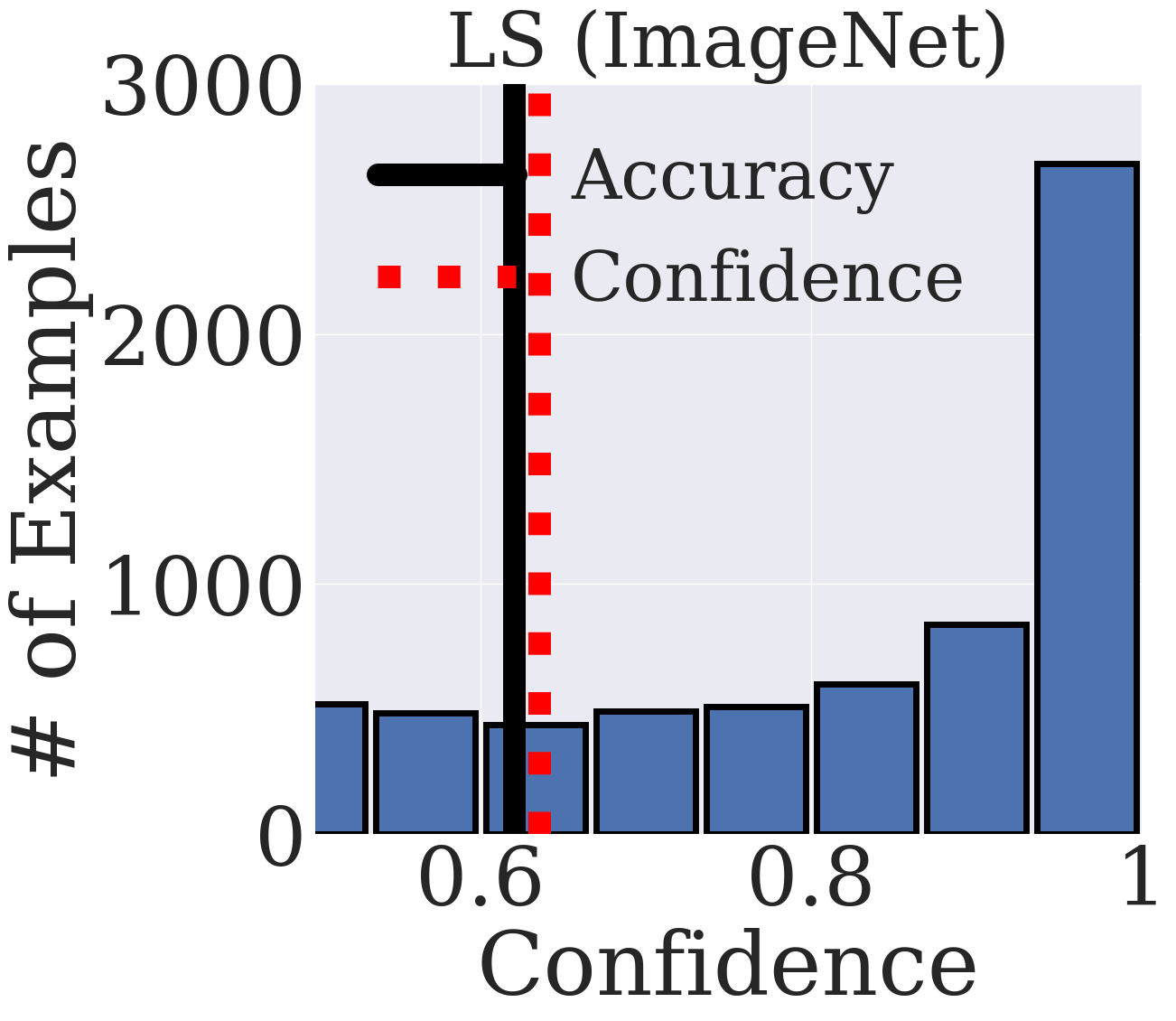}
         \vspace{-0.6cm}
         \caption{}
         \label{fig:over_conf}
     \end{subfigure}
     \begin{subfigure}[b]{0.24\linewidth}
         \centering
         \includegraphics[width=\linewidth]{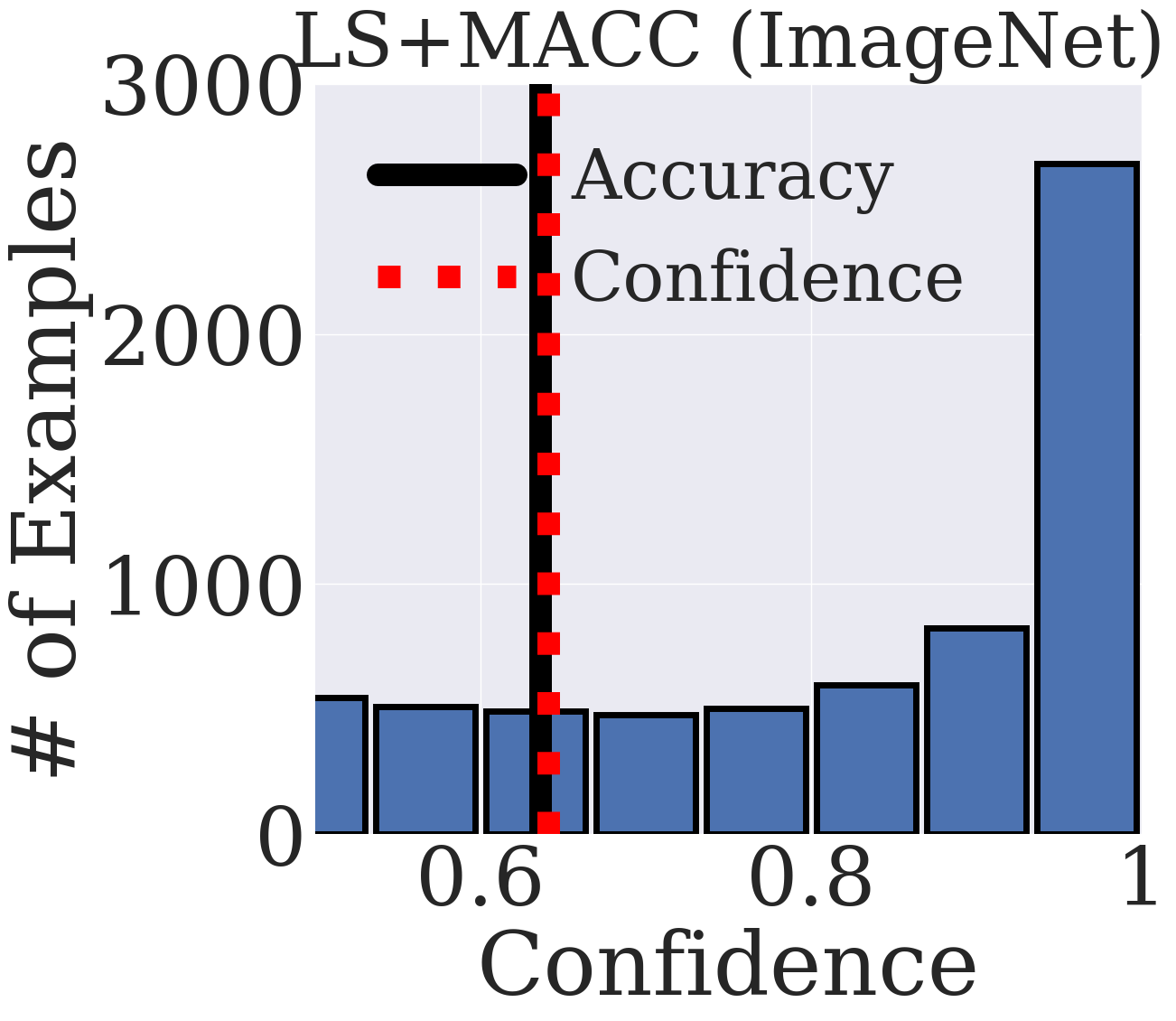}
         \vspace{-0.6cm}
         \caption{}
         \label{fig:over_conf_corrected}
     \end{subfigure}
     \vspace{-0.3cm}
        \caption{Reliability diagrams (a,b,c,d) and confidence histograms (e,f,g,h) of (ResNet) models trained with LS and LS+MACC. (a,b,e,f) show that our method is effective in reducing under-confidence, while (c,d,g,h) reveal that it can reduce over-confidence. We refer to the supplementary for similar plots with NLL and FL.}
        \label{fig:reliability}
\vspace{-0.4cm}
\end{figure}

\noindent\textbf{Mitigating Under/Over-Confidence:}
We plot reliability diagrams to reveal the effectiveness of our method in mitigating under/over-confidence (Fig.~\ref{fig:Teaser_reliability} \&~\ref{fig:reliability} (top)). Furthermore, we plot confidence histograms to illustrate the deviation between the overall confidence (dotted line) and accuracy (solid line) of the predictions (Fig.~\ref{fig:reliability} (bottom)).
Fig.~\ref{fig:under_conf_reliability} \&~\ref{fig:under_conf_corrected_reliability}  show that our method can effectively mitigate the under-confidence of a model trained with LS loss. Fig.~\ref{fig:under_conf} \&~\ref{fig:under_conf_corrected} illustrate that our method notably reduces the gap between the overall confidence and accuracy, thereby mitigating the under-confident behaviour. Likewise, Fig.~\ref{fig:over_conf_reliability},~\ref{fig:over_conf_corrected_reliability},~\ref{fig:over_conf} \&~\ref{fig:over_conf_corrected} display the capability of our method in mitigating the over-confidence of an uncalibrated model.

\begin{figure*}[!htp]
\vspace{-0.5cm}
    \centering
    \begin{subfigure}[b]{0.24\linewidth}\centering
        \includegraphics[width=\linewidth]{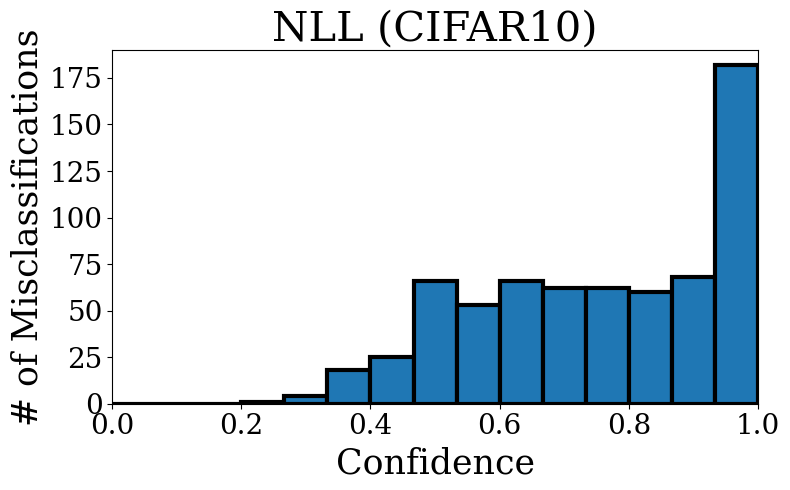}
        \vspace{-0.6cm}
        \caption{}
        \label{fig:nll_cifar10}
    \end{subfigure}
    \begin{subfigure}[b]{0.24\linewidth}
        \centering
        \includegraphics[width=\linewidth]{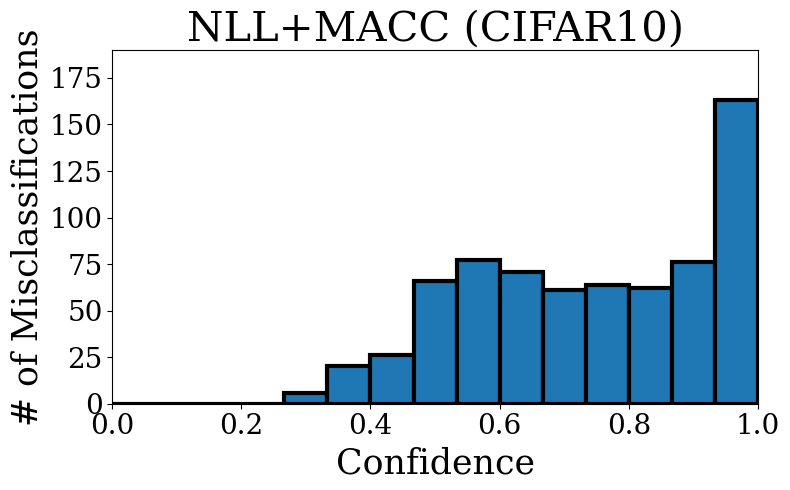}
        \vspace{-0.6cm}
        \caption{}
        \label{fig:nll_mdcu_cifar10}
    \end{subfigure}
    \begin{subfigure}[b]{0.24\linewidth}
        \centering
        \includegraphics[width=\linewidth]{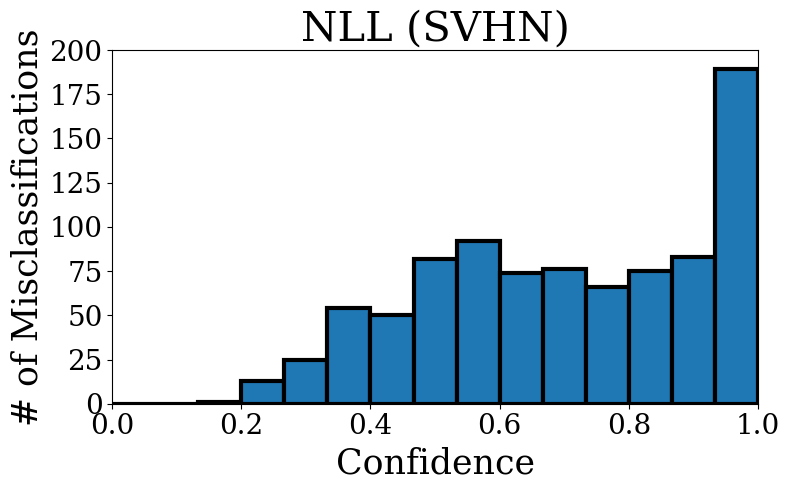}
        \vspace{-0.6cm}
        \caption{}
        \label{fig:fl_cifar10}
    \end{subfigure}
    \begin{subfigure}[b]{0.24\linewidth}
        \centering
        \includegraphics[width=\linewidth]{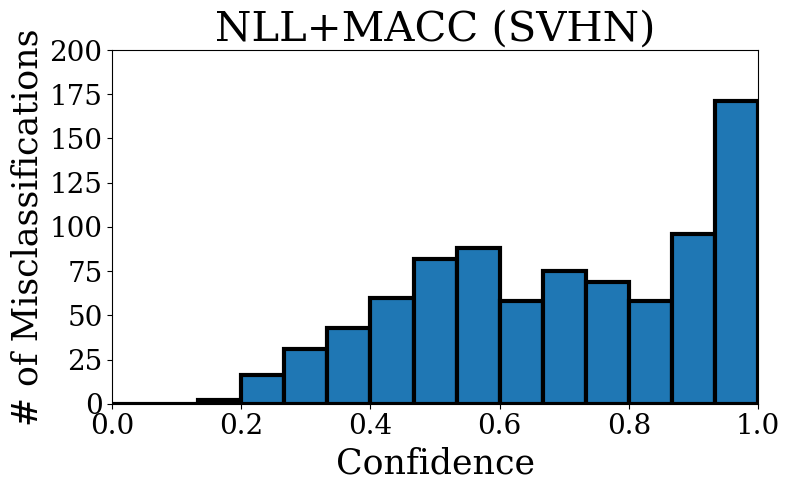}
        \vspace{-0.6cm}
        \caption{}
        \label{fig:fl_mdcu_cifar10}
    \end{subfigure}
    \vspace{-0.35cm}    
    \caption{Confidence histogram of incorrect predictions from CIFAR10 and SVHN (ResNet56).}
    \vspace{-0.6cm}
    \label{fig:misclassified_confidences}
\end{figure*}

\noindent\textbf{Confidence of incorrect predictions:} Fig.~\ref{fig:misclassified_confidences} shows the histogram of confidence values for the incorrect predictions. After adding our auxiliary loss (MACC) to CE loss, the confidence values of the incorrect predictions are decreased (see also Fig.~\ref{fig:Teaser_pacs} \&~\ref{fig:Teaser_in_domain}).

\noindent\textbf{Calibration performance under class imbalance:}
We use SVHN to validate the calibration performance under class imbalance, which has a class imbalance  \mbox{factor} of 2.7 \cite{hebbalaguppe2022stitch}. Both Tables~\ref{tab:comparison_in_domain} and~\ref{tab:classwise_SVHN} show that, compared to competing approaches, MACC not only improves calibration performance over the (whole) dataset, but also displays competitive calibration performance in each class. This is largely because MACC considers whole confidence/certainty vector, which calibrates even the non-predicted classes.



\begin{figure*}[!htp]
     \centering
     \vspace{-0.5cm}
     \begin{subfigure}[b]{0.35\linewidth}
         \centering
\includegraphics[width=\linewidth]{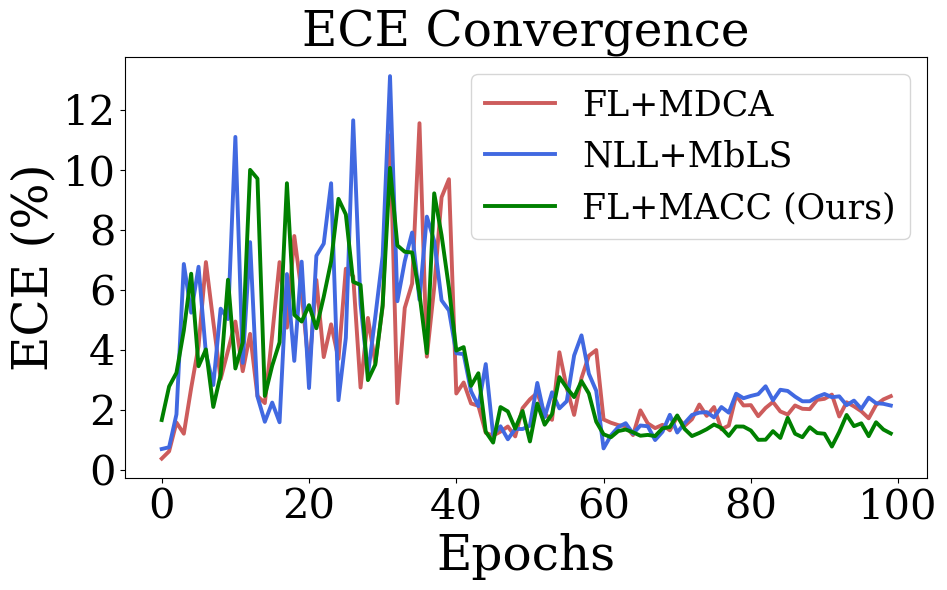}
         \vspace{-0.6cm}
         \caption{}
         \label{fig:sce_convergence}
     \end{subfigure}
     \begin{subfigure}[b]{0.35
     \linewidth}
         \centering
         \includegraphics[width=\linewidth]{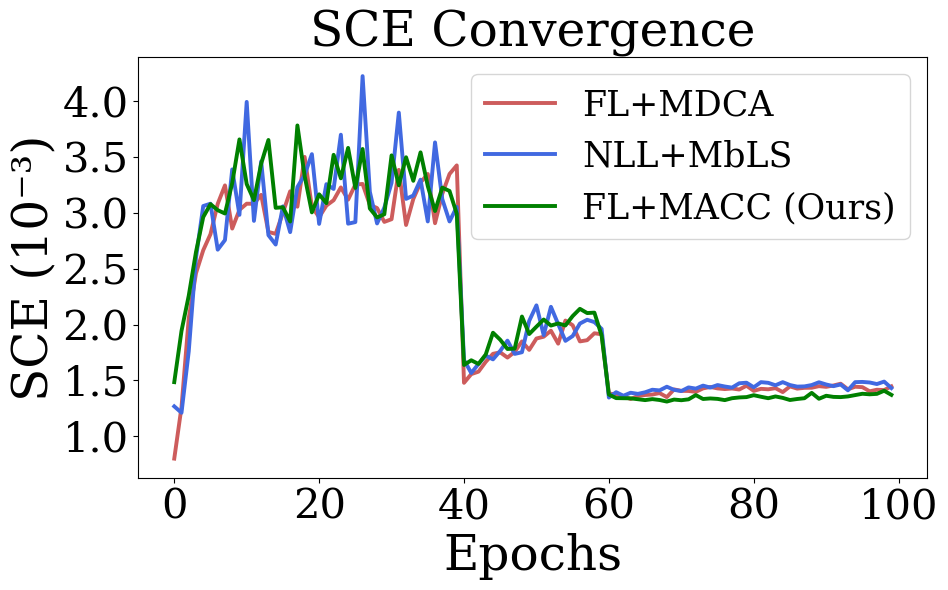}
         \vspace{-0.6cm}
         \caption{}
         \label{fig:ece_convergence}
     \end{subfigure}
        \vspace{-0.4cm}
        \caption{ECE and SCE convergence plot while training ResNet50 on the Tiny-ImageNet for MDCA, MbLS and ours (MACC). We use the learning rate decay factor of 0.1 and 0.02 at epochs 50 and 70, respectively, while for MDCA and MbLS the factor is 0.1.}
        \vspace{-0.7cm}
        \label{fig:convergence}
\end{figure*}

\noindent\textbf{Comparison of ECE and SCE Convergence with SOTA:}
Fig.~\ref{fig:convergence} shows that the proposed loss function MACC is optimizing both ECE and SCE, better than the current SOTA methods of MbLS and MDCA. Although MACC does not explicitly optimize ECE and SCE, it achieves better ECE and SCE convergence. Moreover, compared to others, it consistently decreases both SCE and ECE throughout the evolution of training.

\noindent\textbf{Impact of our model and training settings:}
Table~\ref{tab:dropout_model} shows the performance of the task-specific loss functions and the SOTA calibration losses with the same architecture and training settings as in our loss. i.e., ResNet model with dropout is used and the learning rate at the last stage of the learning rate scheduler is reduced. Upon comparing Table~\ref{tab:dropout_model} with Table~\ref{tab:primary_loss} and Table~\ref{tab:comparison_in_domain}, we note that with our model architecture and learning rate setting, as such, the calibration of competing losses is poor than our loss. So, the effectiveness of our method is not due to the model architecture or some specific training settings but because of our loss formulation.

\begin{table}[!htp]
\vspace{-1cm}
\centering
\tabcolsep=0.1cm
\renewcommand{\arraystretch}{0.80}
\caption{Calibration performance of different losses with our model (ResNet model as in Table~\ref{tab:comparison_in_domain} with dropout) and training settings.}
\resizebox{0.59\textwidth}{!}{
\begin{tabular}{@{}lcccccccc@{}}
\toprule 

\multicolumn{1}{c}{\multirow{2}{*}{\textbf{\footnotesize{Dataset}}}} &
\multicolumn{2}{c}{\textbf{\footnotesize{CE}}} & \multicolumn{2}{c}{\textbf{\footnotesize{FL}}} & \multicolumn{2}{c}{\textbf{\footnotesize{NLL+MbLS}}} & \multicolumn{2}{c}{\textbf{\footnotesize{FL+MDCA}}} \\ \cline{2-9}

& {\footnotesize{SCE}} & {\footnotesize{ECE}} & {\footnotesize{SCE}} & {\footnotesize{ECE}} & {\footnotesize{SCE}} & {\footnotesize{ECE}} & {\footnotesize{SCE}} & {\footnotesize{ECE}} \\ \midrule

\footnotesize{CIFAR10} 
& \footnotesize{$6.43$} & \footnotesize{$2.80$} & \footnotesize{$\underline{3.59}$} & \footnotesize{$\textbf{0.59}$} & \footnotesize{$4.62$} & \footnotesize{$1.35$} & \footnotesize{$\textbf{3.12}$} & \footnotesize{$\underline{0.86}$} \\
\midrule
\footnotesize{CIFAR100} 
& \footnotesize{$2.01$} & \footnotesize{$4.00$} & \footnotesize{$\textbf{1.87}$} & \footnotesize{$\textbf{0.79}$} & \footnotesize{$2.12$} & \footnotesize{$\underline{0.85}$} & \footnotesize{$\underline{2.00}$} & \footnotesize{$1.04$} \\
\midrule
\footnotesize{SVHN} 
& \footnotesize{$1.99$} & \footnotesize{$\textbf{0.27}$} & \footnotesize{$7.30$} & \footnotesize{$3.30$} & \footnotesize{$\underline{1.92}$} & \footnotesize{$\underline{0.34}$} & \footnotesize{$\textbf{1.86}$} & \footnotesize{$0.38$} \\
\cline{1-9}
\bottomrule
\end{tabular}
}
\vspace{-0.2cm}
\label{tab:dropout_model}
\end{table}

\section{Conclusion}
\label{sec:conclusion}

We propose a new train-time calibration method which is based on a novel auxiliary loss term (MACC). Our loss attempts to align the predictive mean confidence with the predictive certainty and is based on the observation that a greater gap between the two translates to higher miscalibration. It is differentiable, operates on minibatches, and acts as a regularizer with other task-specific losses. Extensive experiments on ten challenging datasets show that our loss consistently shows improved calibration performance over the SOTA calibration methods across in-domain and out-of-domain scenarios. 

        \label{fig:timing}

{\small
\bibliographystyle{splncs04}
\bibliography{egbib}
}

\end{document}